\documentclass[letterpaper]{article} 
\usepackage{aaai24}  
\usepackage{times}  
\usepackage{helvet}  
\usepackage{courier}  
\usepackage[hyphens]{url}  
\usepackage{graphicx} 
\usepackage{natbib}  
\usepackage{caption} 
\usepackage{algorithm}
\usepackage{algorithmic}
\usepackage{booktabs}
\usepackage{multirow}
\usepackage{subcaption}
\usepackage{xcolor}
\usepackage{graphicx}
\usepackage{latexsym}
\usepackage{amsmath}
\usepackage{makecell}
\usepackage{fdsymbol}

\usepackage{newfloat}
\usepackage{listings}

\DeclareCaptionStyle{ruled}{labelfont=normalfont,labelsep=colon,strut=off} 
\floatstyle{ruled}
\newfloat{listing}{tb}{lst}{}
\floatname{listing}{Listing}
\title{A Comprehensive Analysis of the Effectiveness of Large Language Models as Automatic Dialogue Evaluators}
\author {
    Chen Zhang\textsuperscript{\rm 1},
    Luis Fernando D’Haro\textsuperscript{\rm 2},
    Yiming Chen\textsuperscript{\rm 1},
    Malu Zhang\textsuperscript{\rm 3}\footnote{corresponding author.},
    Haizhou Li\textsuperscript{\rm 1,4}
}
\affiliations {
    \textsuperscript{\rm 1} National University of Singapore \\
    \textsuperscript{\rm 2} Speech Technology Group - Universidad Politécnica de Madrid, Spain \\
     \textsuperscript{\rm 3} University of Electronic Science and Technology of China \\
    \textsuperscript{\rm 4} The Chinese University of Hong Kong (Shenzhen), China \\
    chen\_zhang@u.nus.edu,
    maluzhang@uestc.edu.cn
}

\usepackage{bibentry}

\begin{document}

\maketitle

\begin{abstract}
Automatic evaluation is an integral aspect of dialogue system research. The traditional reference-based NLG metrics are generally found to be unsuitable for dialogue assessment. Consequently, recent studies have suggested various unique, reference-free neural metrics that better align with human evaluations. Notably among them, large language models (LLMs), particularly the instruction-tuned variants like ChatGPT, are shown to be promising substitutes for human judges. Yet, existing works on utilizing LLMs for automatic dialogue evaluation are limited in their scope in terms of the number of meta-evaluation datasets, mode of evaluation, coverage of LLMs, etc. Hence, it remains inconclusive how effective these LLMs are. To this end, we conduct a comprehensive study on the application of LLMs for automatic dialogue evaluation. Specifically, we analyze the multi-dimensional evaluation capability of 30 recently emerged LLMs at both turn and dialogue levels, using a comprehensive set of 12 meta-evaluation datasets. Additionally, we probe the robustness of the LLMs in handling various adversarial perturbations at both turn and dialogue levels. Finally, we explore how model-level and dimension-level ensembles impact the evaluation performance. All resources are available at \url{https://github.com/e0397123/comp-analysis}.

\end{abstract}

\section{Introduction}

Evaluation remains a persistent challenge in dialogue system research~\citep{mehri-etal-2022-nsf}. At present, human evaluation is regarded as the most reliable method for comprehensively assessing the quality of dialogue. However, because of the considerable expense and lack of reproducibility associated with human evaluations, automatic measures have been proposed to complement human evaluation. The automatic measures can be categorized into two main groups: reference-based and reference-free. Due to the poor alignment of reference-based metrics, such as BLEU~\citep{papineni-etal-2002-bleu}, with human evaluation~\citep{liu-etal-2016-evaluate}, existing studies mainly focus on developing neural-based reference-free evaluators~\citep{yeh-etal-2021-comprehensive}. Although such reference-free evaluators have demonstrated improved correlations with human evaluation over reference-based metrics, they are still far from being the perfect proxy of human judges. Additionally, they exhibit poor generalization to dialogue data that are different from what they are trained on~\citep{zhang-etal-2022-mdd}.

The recent advancement in large language models~\citep{brown2020language,chowdhery2022palm,touvron2023llama} coupled with refined alignment techniques~\citep{wei2022finetuned,ouyang2022training} lead to an extensive array of general-purpose AI assistants that are capable of tackling a broad spectrum of NLP tasks. Harnessing their strong language understanding capability, LLMs, especially the family of instruction-tuned models, offer great promise as effective and generalized automatic dialogue evaluators. Several recent works~\citep{huynh2023understanding,chen2023exploring,liu2023gpteval,lin-chen-2023-llm,fu2023gptscore} report strong correlations of LLMs with human evaluation. Yet, the scope of their assessment is limited: (1) The coverage of LLMs is restricted, with a primary emphasis on proprietary models, such as OpenAI ChatGPT/GPT-4. Lately, there has been an exponential growth of open-source foundation models~\citep{falcon40b,touvron2023llama,workshop2023bloom} and ChatGPT-like LLMs~\citep{alpaca,vicuna2023,chen2023phoenix}, a timely survey is neccessary to examine their effectiveness as automatic dialogue evaluators. (2) The mode of evaluation primarily concentrates on correlation analysis with a limited number of meta-evaluation datasets. We not only conduct a comprehensive correlation analysis on 12 meta-evaluation datasets\footnote{6 turn-level and 6 dialogue-level respectively.} along different evaluation dimensions\footnote{Dimension refers to the quality aspect that is usually assessed in the human evaluation process, such as relevance at the turn level and dialogue coherence at the dialogue level.}, but also probe the robustness of LLMs against adversarial perturbations at both turn and dialogue levels. 

Our work serves as a useful guide for future research on applying LLMs to automatic dialogue evaluation and the contributions are listed as follows:

\begin{itemize}
    \item We conduct a comprehensive analysis of the multi-dimensional evaluation ability of 30 recent LLMs at both turn and dialogue levels. Specifically, we evaluate coherence, engagingness, diversity, informativeness, and overall quality at the dialogue level. For turn-level evaluation, we assess context relevance, understandability, interestingness, specificity, and overall quality.
    
    \item Such a comprehensive assessment is impossible without the availability of large-scale meta-evaluation datasets with annotations. Hence, we complement 12 existing meta-evaluation datasets by providing annotations that were previously unavailable. The datasets together with the full annotations will be made publicly available for researchers and practitioners to benchmark their new evaluation metrics. 
    \item Besides correlation analysis, we introduce a series of adversarial perturbation strategies to reduce the response or dialogue quality along various dimensions. In this way, we can probe the robustness of the LLMs, which has not been explored in existing works.
    \item Lastly, we study the impact of different ensemble strategies on the dialogue evaluation performance, including dimension-level and model-level ensembles.
    
\end{itemize}

\begin{table*}[!ht]
\centering
\resizebox{\linewidth}{!}{
\begin{tabular}{l|c|c|c|c|c|c}
\toprule
\textbf{Turn-Level Datasets} & \textbf{\#Data} & \textbf{\#Utt} & \textbf{Doc Len}  &\textbf{IAA Range} & \textbf{Reused Annotations} & \textbf{Missing Annotations}\\ \midrule
Persona-USR~\shortcite{mehri-eskenazi-2020-usr} & 300 & 9.3 & 98.4 / 12.0 & $0.3\sim0.7$ & rel, int, und, ovr & spe \\
Persona-Zhao~\shortcite{zhao-etal-2020-designing} & 900 & 5.1 & 48.8 / 11.5 & $>0.7$ & ovr & rel, int, und, spe \\
DailyDialog-Zhao~\shortcite{zhao-etal-2020-designing} & 900 & 4.7 & 47.5 / 11.0 & $>0.7$ & rel, ovr  & int, und, spe \\
Topical-USR~\shortcite{mehri-eskenazi-2020-usr} & 360 & 11.2 & 236.3 / 22.4 & $0.5\sim0.7$ & rel, int,  und, ovr & spe \\
FED-Turn~\shortcite{mehri-eskenazi-2020-unsupervised}  & 375 & 10.4 & 87.3 / 13.3 & $0.5\sim0.8$ & rel, int, spe, und, ovr & - \\
ConTurE-Turn~\shortcite{ghazarian-etal-2022-wrong} & 1066 & 3.8 & 21.7 / 11.0 & $\sim0.3$ & ovr & rel, int, und, spe \\
\midrule
\textbf{Dialogue-Level Datasets} & \textbf{\#Data} & \textbf{\#Utt} & \textbf{Doc Len}  &\textbf{IAA Range} & \textbf{Reused Annotations} & \textbf{Missing Annotations} \\ \midrule
IEval-Dial~\shortcite{svikhnushina-etal-2022-ieval} & 500 & 6.0 & 74.4 & - & - & coh, eng, inf, div, ovr \\
Persona-See~\shortcite{see-etal-2019-makes} & 500 & 12.0 & 91.2 & - & -   & coh, eng, inf, div, ovr \\
Reliable-Dial~\shortcite{ji-etal-2022-achieving} & 500 &  21.2 & 178.1 & - & - & coh, eng, inf, div, ovr \\
ConTurE-Dial~\shortcite{mehri-etal-2022-interactive} & 119 & 17.9 & 153.9 & - &  - & coh, eng, inf, div, ovr \\
FED-Dial~\shortcite{mehri-eskenazi-2020-unsupervised} & 125 & 12.7 & 116.8 & $0.7\sim0.8$ & coh, eng, inf, div, ovr & - \\
Human-Eval~\shortcite{smith-etal-2022-human} & 286 & 12.0 & 139.2 & - & - & coh, eng, inf, div, ovr \\
\bottomrule
\end{tabular}
}
\caption{Details of the meta-evaluation datasets. The ``Reused" columns indicate the dimensions with available human-annotated scores. The ``Missing" column denotes the dimensions that need to be annotated by us. The doc length is the average \#words per context/response for turn-level datasets and dialogue for dialogue-level datasets. The IAA range shows the range of inter-annotator agreements of available human annotations. \#Utt refers to the number of context utterances and dialogue utterances for the turn-level and dialogue-level datasets respectively.}\label{tab:dataset-details}
\end{table*}

\section{Preliminaries}

\subsection{Meta-Evaluation}
\label{subsec:metaeval-datasets}

\subsubsection{Datasets}
We adopt 12 meta-evaluation datasets in our analysis comprising 6 at the turn level and another 6 at the dialogue level. Table~\ref{tab:dataset-details} summarizes the dataset details. For both turn-level and dialogue-level analysis, we evaluate five different quality aspects/dimensions respectively. At the turn level, we assess context relevance (rel), understandability (und), specificity (spe), interestingness (int), and overall response quality (ovr) while at the dialogue level, we evaluate coherence (coh), engagingness (eng), informativeness (inf), diversity (div), and overall dialogue quality (ovr).

\begin{figure}[!ht]
    \centering
    \resizebox{0.9\linewidth}{!}{
    \includegraphics{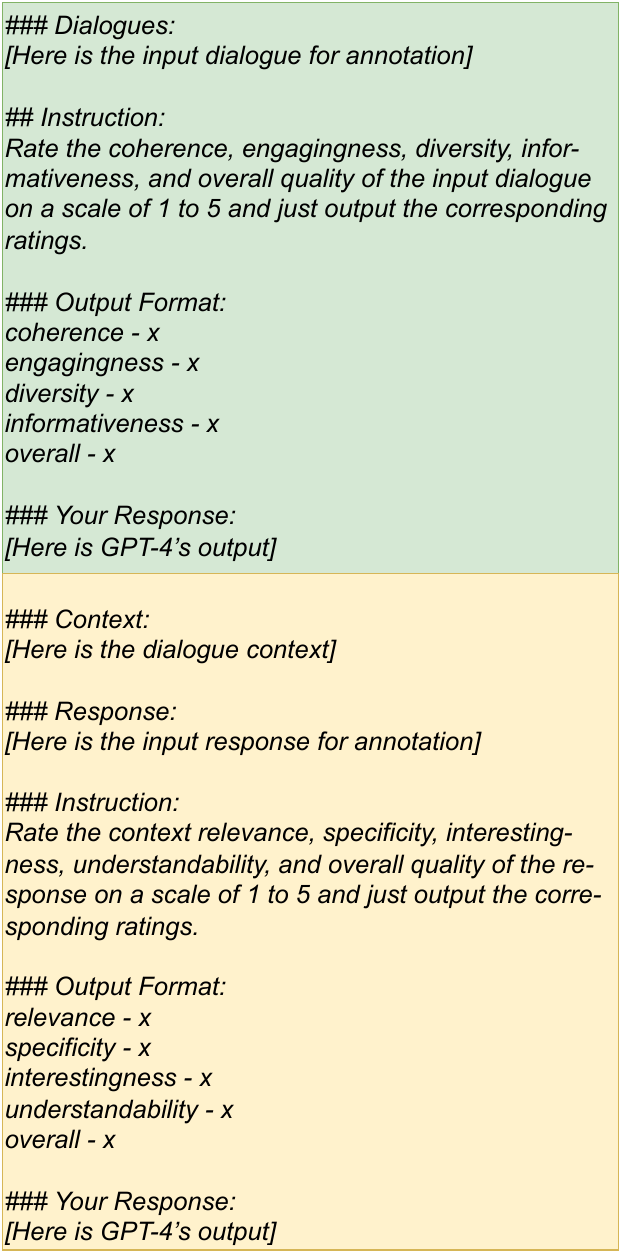}}
    \caption{The instruction template for prompting GPT-4 to annotate both dialogue-level (top) and turn-level (bottom) data. For our meta-evaluation of the proprietary models including ChatGPT and Palm-2 Bison, we also use this instruction template.}
    \label{fig:gpt-prompt-template}
\end{figure}

\subsubsection{Fill Up Missing Annotations With GPT-4} 
To save costs and speed up the annotation process, we perform the necessary annotations with GPT-4 instead of using crowd-workers. The motivation is that existing works show that GPT-4 can achieve human-level judgment while being more scalable and less expensive~\citep{zheng2023judging,gilardi2023chatgpt,liu2023gpteval}. It serves as a supplementary tool to human evaluation, especially when considering the high costs and the reproducibility issues of human annotations in open-ended generation tasks~\citep{karpinska-etal-2021-perils}. Note that GPT-4 annotations may contain biases and future works can explore automatic and manual ways to mitigate the biases.

Figure~\ref{fig:gpt-prompt-template} illustrates how we prompt GPT-4 to perform the annotation task. When calling the GPT-4 API, we set the temperature and top\_p to 0.7 and 0.95 respectively. The annotation process is repeated five times to mimic the scenario of having multiple crowd workers annotate each data instance. The inter-annotator or inter-round agreement for GPT-4 is derived by averaging the pairwise Pearson correlations between the annotation scores from any two annotation rounds. In general, a good inter-annotator or inter-round agreement exceeding 0.65 is observed when annotating the missing dimensions of different datasets.

\subsubsection{Meta-Evaluation Metrics}
\label{subsec:metaeval-metrics}
The reliability of LLMs as automatic evaluators is assessed by computing how well their evaluation scores ($s^{dim}_{llm}$) correlate with the corresponding human ($s^{dim}_{human}$) or GPT-4 judgment ($s^{dim}_{gpt4}$) with a correlation function $g$ for a specific dimension. $s^{dim}_{human}$ is derived by averaging the annotations of multiple annotators while $s^{dim}_{gpt4}$ is obtained by averaging the annotations from the multiple annotation rounds. We adopt the commonly-used Pearson ($\rho$) measure as $g$.

\subsection{Large Language Models}

30 LLMs comprising 28 open-source and 2 proprietary LLMs are examined. The proprietary LLMs are OpenAI ChatGPT\footnote{We use the 2023-03-15-preview version on Azure.} (gpt-3.5-turbo) and Google Palm-2 Bison (text-bison-001). The performance of OpenAI GPT-4 on human-annotated data is also reported. The 28 open-source LLMs can be grouped into two categories, the vanilla foundation models and the instruction-tuned models. The foundation models include different variants of Meta LLaMA~\citep{touvron2023llama,touvron2023llama2}, SalesForce XGen\footnote{\url{https://blog.salesforceairesearch.com/xgen/}}, TII-UAE Falcon~\citep{falcon40b}, MosaicML MPT\footnote{\url{https://huggingface.co/mosaicml/mpt-7b}}, OpenLLaMA\footnote{\url{https://github.com/openlm-research/open_llama}}, Pythia~\citep{biderman2023pythia}, and BLOOM~\citep{workshop2023bloom}. The instruction-tuned models are mainly derivatives of the aforementioned vanilla foundation models, such as Alpaca~\citep{alpaca}, Vicuna~\cite{vicuna2023}, Tulu~\citep{wang2023far}, and Chimera~\citep{chen2023phoenix}. They are finetuned to mimic the abilities of proprietary LLMs, such as ChatGPT and GPT-4. Alignment techniques, such as instruction-based supervised finetuning (SFT) and reinforcement learning from human feedback (RLHF), are applied to align these models with humans' general task-solving abilities. We refer interested readers to their respective technical reports for the details of these LLMs.


    
    


    
    


\begin{figure}[!ht]
    \centering
    \resizebox{0.85\linewidth}{!}{
    \includegraphics{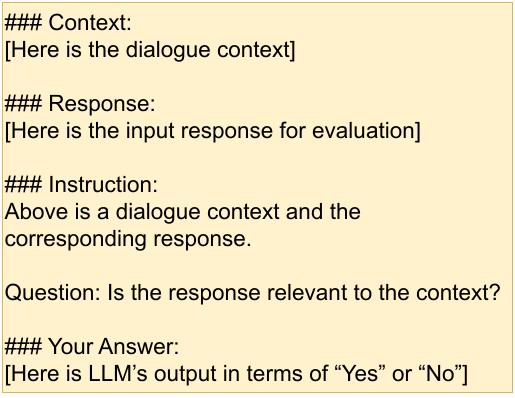}}
    \caption{An example for prompting open-source LLMs to evaluate the contextual relevance of the input response.}
    \label{fig:llm-ins-template}
\end{figure}

\subsection{Dialogue Evaluation with LLMs}

As we cannot obtain the output probabilities of the proprietary models at the time of paper preparation, we adopt the explicit scoring procedure to directly prompt them to produce multi-dimensional ratings of dialogues or responses. ChatGPT, Palm-2 Bison, and GPT-4 share the same instruction template outlined in Figure~\ref{fig:gpt-prompt-template}. Due to their strong instruction-following abilities, we can easily extract the scores with matching heuristics. The rare erroneous cases are manually fixed.  We repeat the scoring process of each proprietary model five times and apply the average score of the five runs as the corresponding $s^{dim}_{llm}$. 

We do not apply the same approach for obtaining ratings from other open-source LLMs because their instruction-following abilities are weaker than the proprietary models. Sometimes, their generation becomes intractable. Instead, we follow~\citet{gupta-etal-2022-instructdial} by applying an implicit scoring procedure. More specifically, given an instruction prompt input, we concentrate on the output probabilities related to the label words "Yes" and "No" as generated by the LLM. Then, we normalize the probability of ``Yes" as $P(Yes) = P(Yes)/(P(Yes) + P(No))$ and $P(Yes)$ serves as the corresponding $s^{dim}_{llm}$. Greedy decoding is applied such that the generation process of the LLMs is deterministic. Figure~\ref{fig:llm-ins-template} showcases an example to prompt open-source LLMs to evaluate response relevance\footnote{All dimension-specific questions for prompting the open-source LLMs can be found in the Appendix.}. For different LLMs, we adapt the instruction template to match that used in their instruction-tuning process.

\section{Multi-Dimensional Correlation Analysis}
\label{sec:multi-eval}

\begin{table*}[!t]	
\centering
\resizebox{\linewidth}{!}{
\begin{tabular}{c|cccccc|cccccc}
\toprule

 & \multicolumn{6}{c}{\textbf{Turn-Level}} & \multicolumn{6}{c}{\textbf{Dialogue-Level}}\\
\midrule
\textbf{Models} & \textbf{Rel} & \textbf{Spe} & \textbf{Int} & \textbf{Und}& \textbf{Ovr} & \textbf{Avg} & \textbf{Coh} & \textbf{Eng} & \textbf{Div} & \textbf{Inf}& \textbf{Ovr} & \textbf{Avg} \\ \midrule
LLaMA-7B &0.114 &0.056 & 0.073& 0.113&0.018& 0.075 &0.341 &0.142 &0.049& 0.166& 0.109& 0.161 \\ 
 LLaMA-13B &0.442 & 0.240& 0.333& 0.348& \underline{0.402} & 0.353& 0.522 &0.425 &0.237&0.324& 0.404&0.382 \\ 
LLaMA-2-7B & 0.241& 0.187&0.178&0.134 & 0.259& 0.200&0.225 & 0.227&0.044 &0.261 & 0.292 & 0.210\\ 
LLaMA-2-13B &0.303 &0.122 &0.161&0.269 & 0.225& 0.216& 0.318& 0.362& -0.068& 0.343& 0.417&0.274 \\ 
XGen-8K-7B & 0.214& 0.152& 0.173& 0.119&0.218 &0.175 &0.419 & 0.431& 0.325&0.387 & 0.456&0.404  \\ 
Falcon-7B & 0.133&0.061 &0.174& 0.101& 0.104& 0.114& 0.366& 0.476&0.329 &0.341 & 0.413&0.385  \\ 
MPT-8K-7B & 0.209&0.048 &0.123& 0.197&0.172 &0.150 &0.126 &0.221 &0.112 & 0.187& 0.088&0.147 \\ 
OpenLLaMA-7B & 0.158&0.008 &0.105& 0.154& 0.184& 0.122& 0.402&0.452 &0.270 & 0.336& 0.429& 0.378\\ 
OpenLLaMA-13B &0.231 &0.112 &0.175& 0.184& 0.226&0.186 &0.425 & 0.506& 0.239& 0.342&0.471 & 0.397\\  
Pythia-7B &0.145 & 0.097& 0.083& 0.106& 0.146& 0.115&0.079 &0.204 & 0.026& 0.094&0.220 & 0.124\\ 
BLOOM-7B & 0.082&0.021 &0.124& 0.099& 0.122& 0.089&0.224 & 0.271& 0.137& 0.217&0.277 & 0.225\\ \midrule
LLaMA-2-Chat-7B &0.434 &0.191 &0.243&0.251 & 0.240& 0.272& 0.534&0.485 & 0.375& 0.447&0.498&0.468 \\ 
LLaMA-2-Chat-13B &\underline{0.559} & 0.353& 0.207& 0.320& 0.380&0.364& \underline{0.644} & 0.610& 0.228& 0.514& \underline{0.613}&0.522\\ 
Alpaca-7B &0.430 & 0.293&0.253& 0.267& 0.386&0.326 & \underline{0.586} & \underline{0.624} & 0.372&0.477 & 0.566 & 0.525\\ 
Vicuna-7B & 0.368&0.096 & 0.221& 0.100& 0.219& 0.201&0.490 &0.468 &0.279 & 0.488& 0.482&0.441\\
Vicuna-13B& 0.400& 0.318&0.229& 0.224& 0.309&0.296 & 0.515&0.420 & 0.306& 0.419& 0.417 & 0.415\\ 
Falcon-Ins-7B& 0.272& 0.152& 0.293& 0.179&0.246 & 0.228& 0.504&0.513 &0.375 & 0.448&0.500 & 0.468 \\ 
Tulu-13B & \underline{0.585}& \underline{0.427}&0.369 & \underline{0.350}& \underline{0.488}& \underline{0.444} &\underline{0.659} & \underline{0.661} & 0.326& \underline{0.518}& \underline{0.657} & \underline{0.564}\\ 
Chimera-7B &0.489 &0.276 & \underline{0.373} &0.309 & 0.368&0.363 &0.563 & 0.599&\underline{0.439} & \underline{0.525}& 0.607 & 0.547\\ 
Chimera-13B &\underline{0.547} & \underline{0.449} & \underline{0.377}& \underline{0.366} & \underline{0.404}& \underline{0.428}&0.582 &\underline{0.671} & \underline{0.432} & \underline{0.585}&0.563 & \underline{0.567} \\ 
Phoenix-7B &0.314 & 0.101& 0.291& 0.258&0.234 &0.240 &0.480 &0.493 &0.146 & 0.334&0.416 &  0.374\\ 
Oasst-sft-Pythia-12B &0.144 &0.028 & 0.203&0.132 &0.110 & 0.123& 0.386& 0.358&0.236 & 0.346&0.423 &  0.350\\ 
Baize-v2-13B & 0.477& 0.350& 0.333& \underline{0.353} &0.337 & \underline{0.370} &0.568 & 0.544& 0.397& 0.482& 0.469 & 0.492\\ 
Dolly-v2-12B &0.030 &-0.009 & 0.061&-0.004 &0.020 &0.020 &0.182 & 0.238&0.071 &0.139 &0.105 & 0.147 \\ 
MPT-8K-7B-Instruct &0.139 &0.092 & 0.176&0.095 &0.099 &0.120 & 0.321& 0.352& 0.316&0.308 & 0.315 & 0.322\\ 
XGen-8K-7B-Inst &0.272 &0.293 & 0.267& 0.145& 0.265& 0.248& 0.502&0.515 & 0.308& 0.417& 0.506 & 0.450\\ 
ChatGLM-v2-6B &0.368 & 0.267& 0.191&0.181 &0.184 &0.238 &0.214 & 0.236& 0.262& 0.248& 0.359 &0.264 \\ 
WizardLM-13B-V1.2 &0.463 & \underline{0.422} & \underline{0.390}& 0.245&0.314 & 0.367& 0.572& 0.580& \underline{0.455}& 0.513&\underline{0.632} & \underline{0.550} \\ \midrule
Palm-2 (text-bison-001)&\underline{0.666} & \underline{0.563} &\underline{0.454}& \underline{0.422} & \underline{0.601}& \underline{0.541}&\underline{0.649} &\underline{0.674} & \underline{0.473}& \underline{0.557}& \underline{0.674} & \underline{0.605}\\ 
ChatGPT (gpt-3.5-turbo) & \underline{0.595}&\underline{0.578} &\underline{0.518}& \underline{0.536} & \underline{0.542} & \underline{0.554} &\underline{0.724}& \underline{0.705}& \underline{0.516}&\underline{0.568} & \underline{0.707} & \underline{0.644}\\ 
\bottomrule
\end{tabular}}
\caption{Dimension-specific Pearson correlation scores of different LLMs on fully-annotated data. The top-five scores in each dimension are underlined. 7B refers to 7 billion number of parameters.}
\label{tab:full-corr-results}
\end{table*}

We report dimension-wise Pearson correlation scores averaged across either the turn-level or the dialogue-level datasets\footnote{For example, a model's performance on a dimension at the turn level is derived by averaging the correlations obtained across the six turn-level meta-evaluation datasets w.r.t. that dimension.} in Table~\ref{tab:full-corr-results}. 

\subsubsection{Proprietary vs Open-Source Models}
It can be observed that ChatGPT and Palm-2 Bison are among the top 5 models across all the dimensions at both turn and dialogue levels. On average, they ranked first and second respectively. Out of the 28 open-source models available, Tulu-13B~\citep{wang2023far}, Chimera-inst-chat-13B~\citep{chen2023phoenix}, and Baize-v2-13B~\citep{xu2023baize} are distinguished as the top three performers at the turn-level. When it comes to dialogue-level performance, the top three models are Chimera-inst-chat-13B, Tulu-13B, and WizardLM-13B-V1.2~\citep{xu2023wizardlm}. A significant gap can be observed between the proprietary models and the best open-source model. For example, at the turn level, ChatGPT outperforms Tulu-13B by 11\% on average while at the dialogue level, ChatGPT outperforms Chimera-inst-chat-13B by 7.7\%. The observations suggest the importance of the model scale and the quality of instruction data. The proprietary models are much larger than other open-source LLMs. They are trained on more sophisticated human-annotated instruction data while the open-source models are mainly instruction-tuned on data distilled from proprietary models. 

\subsubsection{Instruction-Tuned vs Vanilla Models}
We can also observe that instruction-tuned variants generally outperform their corresponding vanilla backbone models. For instance, Alpaca-7B surpasses LLaMA-7B by a significant 25.1\% on average at the turn level and 36.4\% on average at the dialogue level. Similarly, Tulu-13B beats LLaMA-13B by a notable 9.1\% and 18.2\% on average. The observations showcase that alignment techniques, such as instruction-based supervised finetuning, can greatly enhance the dialogue understanding capabilities of LLMs, thereby making them more useful for automatic dialogue evaluation. 

\subsubsection{LLaMA vs Other Open-Source Families}
Among 7B vanilla models, LLaMA-2-7B tops the list at the turn level with an average of 0.200, while XGen-8K-7B leads at the dialogue level with an average of 0.404. For the 13B vanilla LLMs, LLaMA-13B stands out at the turn level with a 0.353 score on average, and OpenLLaMA-13 takes the lead at the dialogue level with an average score of 0.397. We can also observe a large performance variation among the vanilla models. For example, the gap between LLaMA-2-13B and BLOOM-7B is more than 25\% at the turn level. This can be attributed to the differences in their model sizes, pretraining corpora, and training strategies. 

In the 7B instruction-tuned category, Alpaca-7B and Chimera-inst-chat-7B stand out as the top two performers across both turn and dialogue levels, both of which are derivatives of LLaMA-7B. Given that LLaMA-13B serves as a strong foundational model for automated dialogue evaluation, it's understandable why its instruction-tuned variants, including Tulu-13B, Chimera-inst-chat-13B, Baize-v2-13B, and WizardLM-13B-V1.2, are among the best open-source models. With the observations, we can conclude that by far, models in the LLaMA family are stronger automatic dialogue evaluators than other open-source LLMs.

\subsubsection{Impact of Instruction Data}
Even among the models with the same number of parameters and the same backbone, the performance varies greatly. For instance, Tulu-13B outperforms Vicuna-13B by 0.148 and 0.149 at turn and dialogue levels respectively in terms of the average Pearson correlations. The common attribute shared by Tulu-13B and Chimera-inst-chat-13B is that they use a more diverse set of instruction data than other models. Their instruction data come from different sources. For instance, the training data of Tulu consist of collections of NLP datasets with human-written instructions like FLAN V2~\citep{chung2022scaling} and CoT~\citep{wei2022chain}, collections of human-written instruction data from scratch, such as Dolly\footnote{\url{https://www.databricks.com/blog/2023/04/12/dolly-first-open-commercially-viable-instruction-tuned-llm}}, and data mixture that is distilled from proprietary models like text-Davinci-003, ChatGPT, and GPT-4. Thus, it's evident that the diversity and quality of instruction data can significantly influence the model performance, emphasizing the importance of sourcing diverse datasets that better align with the task when finetuning our LLM-based evaluators.

\subsubsection{Performance Across Dimensions}
Generally, the majority of models excel in areas like relevance, coherence, engagement, and overall quality. However, their performance diminishes slightly when assessing interestingness and understandability at the turn level, and diversity at the dialogue level. Future explorations into leveraging LLMs for automatic dialogue evaluation might benefit from designing objectives that target these specific dimensions. 

\subsubsection{Performance of GPT-4}
\label{subsec:gpt4-performance}

The turn-level and dialogue-level evaluation abilities of GPT-4 are compared to the respective top-5 LLMs.  The results are presented in Table~\ref{tab:gpt4-results}. Note that Table~\ref{tab:gpt4-results} cannot be directly matched to Table~\ref{tab:full-corr-results} because we only use the human-annotated data for correlation computation here while in the previous section, we conduct correlation analysis on the full data, which contain both human annotations and GPT-4 annotations. From Table~\ref{tab:gpt4-results}, we can see that GPT-4 achieves the best correlations with human evaluation in almost all the dimensions, except for specificity at the turn level and diversity at the dialogue level. It performs exceptionally well for relevance, coherence, and overall quality. On average, GPT-4 outperforms the second-best LLM (ChatGPT) by a large absolute margin of 0.085 and 0.042 at turn and dialogue levels respectively. The observations justify our using GPT-4 to complete the missing annotations in the datasets. However, we should note that even the powerful GPT-4 model cannot reach perfect correlations ($>0.8$) on average suggesting that automatic dialogue evaluation remains an open problem. Future research should continue enhancing the conversation understanding capabilities of the language models.

\begin{table}[!t]	
\centering
\Large
\resizebox{\linewidth}{!}{
\begin{tabular}{c|ccccc|c}
\toprule
\multicolumn{7}{c}{\textbf{Turn-Level}} \\ \midrule
 & \textbf{Rel} & \textbf{Spe} & \textbf{Int} & \textbf{Und}& \textbf{Ovr} & \textbf{Avg}  \\ \midrule
Baize&0.449&0.147&0.302&0.290&0.337&0.305 \\ 
Tulu&0.544&0.193&0.254&0.324&0.488&0.361 \\ 
Chimera&0.507&0.234&0.312&0.316&0.404&0.354 \\ 
Palm-2 &0.616&0.317&0.343&0.384&0.601&0.452 \\ 
ChatGPT &0.576&\textbf{0.408}&0.446&0.424&0.542&0.479 \\ 
GPT-4 &\textbf{0.704}&0.342&\textbf{0.538}&\textbf{0.558}&\textbf{0.677}& \textbf{0.564}\\
\midrule
\multicolumn{7}{c}{\textbf{Dialogue-Level}}\\ \midrule
& \textbf{Coh} & \textbf{Eng} & \textbf{Div} & \textbf{Inf}& \textbf{Ovr} & \textbf{Avg}  \\ \midrule
Tulu&0.668&0.629&0.414&0.584&0.681& 0.595\\ 
Chimera&0.595&0.628&0.507&0.609&0.525&0.573 \\ 
 WizardLM &0.536&0.522&0.477&0.540&0.605& 0.536\\
Palm-2&0.584 &0.633&0.550&0.604&0.614& 0.597\\ 
ChatGPT &0.650 & 0.647 & \textbf{0.551}& 0.570 & 0.715 & 0.627\\ 
GPT-4 & \textbf{0.760} & \textbf{0.689} & 0.534 & \textbf{0.620} & \textbf{0.744} & \textbf{0.669} \\
\bottomrule
\end{tabular}}
\caption{Dimension-specific Pearson correlation scores of GPT-4 vs other strong LLMs on human-annotated data. The best score in each dimension is highlighted in bold. All the open-source LLMs are the 13B variants.}
\label{tab:gpt4-results}
\end{table}

\subsubsection{Deviation Between GPT-4 and Human Preferences} 

We analyze how much the assessment of the LLMs based on GPT-4 annotations deviates from that based on human annotations. Specifically, we compare the ranking results of the 32 LLMs evaluated by GPT-4 vs those assessed by human annotations. The deviation is quantified by the Spearman correlation between the two ranking lists. A greater Spearman correlation indicates smaller deviations in the model rankings. We perform the analysis on the FED dataset~\citep{mehri-eskenazi-2020-unsupervised}, which contains human annotations for all the dimensions. As shown in Table~\ref{tab:deviation}, the results reveal minimal deviation of GPT-4's evaluation from human evaluation ($>$0.85 agreements) in all dimensions except response specificity. 

\begin{table}[!ht]	
\centering
\resizebox{0.9\linewidth}{!}{
\begin{tabular}{c|cccccc}
\toprule
 & \textbf{Rel} & \textbf{Spe} & \textbf{Int} & \textbf{Und}& \textbf{Ovr}  \\ \midrule
FED-Turn & 0.950 & 0.551 & 0.887 & 0.869 & 0.971 \\ \midrule
& \textbf{Coh} & \textbf{Eng} & \textbf{Div} & \textbf{Inf}& \textbf{Ovr} \\ \midrule
FED-Dial & 0.861 & 0.900 & 0.883 & 0.869 & 0.945 \\ 
\bottomrule
\end{tabular}}
\caption{Spearman of model rankings evaluated by GPT-4 ratings vs evaluated by human ratings.}
\label{tab:deviation}
\end{table}

\section{Ensemble Analysis}
\label{sec:ensemble}

In this section, we delve into two straightforward ensemble strategies. The first approach involves averaging the scores of each LLM assigned to different dimensions\footnote{At turn level, the relevance, specificity, understandability, and interestingness scores are averaged to derive an overall score while at the dialogue level, the coherence, engagingness, diversity, and informativeness scores are averaged.} to see if this average provides a stronger correlation with overall human evaluations than directly prompting the LLM to assess the overall quality. We refer to this method as the "dimension-wise ensemble". The second strategy entails averaging scores from multiple LLMs for a given dimension, allowing us to determine if this method can match the performance of the stronger proprietary LLMs. We call this approach the "model-wise ensemble".  


\subsubsection{Dimension-Wise Ensemble} 

\begin{table}[!ht]
\centering
\resizebox{\linewidth}{!}{
\begin{tabular}{@{}l|ccccc@{}}
\toprule
& \multicolumn{2}{c}{\textbf{Turn Level}}   &     & \multicolumn{2}{c}{\textbf{Dialogue Level}}
\\\cmidrule{2-3} \cmidrule{5-6}
\textbf{Model}    &  \textbf{Ensemble} & \textbf{Direct} &       & \textbf{Ensemble} & \textbf{Direct} \\ \midrule
Tulu   & 0.492 & 0.488   &     & 0.656 & 0.657     \\
Chimera  & 0.476 &  0.404  &    &  0.680 & 0.563  
\\ 
Baize    & 0.417 &  0.337  &    & 0.578 & 0.469 
\\ 
WizardLM   & 0.408 & 0.314  &    & 0.598 & 0.632 
\\ 
Palm-2  & 0.602 &  0.601  &    & 0.693 & 0.674  
\\ 
ChatGPT  & 0.564 &  0.542  &    & 0.709 & 0.707  
\\ 
\bottomrule
\end{tabular}
}
\caption{Dimension-wise ensemble results of difference models for the overall quality evaluation. All the open-source LLMs are the 13B variants.}
\label{tab:dimension-ensemble-results}
\end{table}

We limit the analysis to Tulu-13B, Chimera-inst-chat-13B, Baize-v2-13B, WizardLM-13B-V1.2, Palm-2 Bison, and ChatGPT, which have strong multi-dimensional evaluation performance. Table~\ref{tab:dimension-ensemble-results} presents performance comparisons between the dimension-wise ensemble and the direct prompting approaches for evaluating the overall response or dialogue quality of each model. We can observe that the ensemble approach yields strong correlations with overall human judgments in general. Especially for Chimera-inst-chat-13B and Baize-v2-13B at the dialogue level, the ensemble approach provides gains of more than 10\% than direct prompting. We also observe that the scores assigned by Tulu-13B, ChatGPT, and Palm-2 Bison to different dimensions are highly similar while Chimera-inst-chat-13B and Baize-v2-13B provide more diverse scores when evaluating different dimensions. This may explain why the ensemble of different dimension-specific scores of Chimera-inst-chat-13B and Baize-v2-13B leads to more significant improvements than other LLMs.

\subsubsection{Model-Wise Ensemble}

\begin{table}	
\centering
\small
\resizebox{0.9\linewidth}{!}{
\begin{tabular}{@{}c|ccc@{}}
\toprule
\multicolumn{4}{c}{\textbf{Turn Level}} \\ \midrule
\textbf{Dimensions}  & \textbf{Palm-2 Bison} & \textbf{ChatGPT} & \textbf{Ensemble} \\ \midrule
Rel & 0.666 & 0.595 & 0.632 \\
Int &0.454 & 0.518 & 0.465\\
Und &0.422 & 0.536 & 0.407 \\
Spe & 0.563&0.578 & 0.487\\
Ovr &0.601 & 0.542&0.491 \\ \midrule
Average &0.541 & 0.554& 0.496 \\ \midrule
\multicolumn{4}{c}{\textbf{Dialogue Level}} \\ \midrule
Coh & 0.649& 0.724& 0.700 \\
Eng & 0.674&0.705 &0.725 \\
Div & 0.473& 0.516& 0.492\\
Inf &0.557 &0.568 & 0.610 \\
Ovr &0.674 &0.707 & 0.686\\ \midrule
Average &0.605 &0.644 & 0.643 \\
\bottomrule
\end{tabular}}
\caption{Performance of the ensemble vs proprietary models for each dimension at the turn and dialogue levels.}
\label{tab:model-ensemble-results}
\end{table}

For the model-wise ensemble, we average the scores of the top 3 open-source models for each dimension as indicated in Table~\ref{tab:full-corr-results}. We can observe that the ensemble achieves comparable performance to ChatGPT and outperforms Palm-2 Bison at the dialogue level. At the turn level, the ensemble's performance is worse than that of both ChatGPT and Palm-2 Bison for dimensions other than relevance and interestingness. The simple ensemble showcases the potential benefits of combining multiple models to boost evaluation performance. Future research might delve deeper into optimal ways of ensembling, such as how to best combine models, which models to include in the ensemble, and how to weigh individual model outputs. 

\begin{table}[!ht]	
\centering
\small
\resizebox{0.9\linewidth}{!}{
\begin{tabular}{@{}l|ccccc@{}}
\toprule
\textbf{} & \textbf{Level} & \textbf{Source} & \textbf{\#Data} & \textbf{Dims} &  $\mathbf{\theta}$ \\ \midrule
RR & Turn & DD \& PC & 1894 & Rel & 0.3  \\
RP & Turn & DD \& PC & 2919 & Rel & 0.2  \\
RNE & Turn & DD \& PC & 712 & Rel & 0.2 \\
Con & Turn & DD \& PC & 1094 & Rel & 0.2  \\ \midrule
Rep & Turn & DD \& PC & 1993 & Und & 0.2 \\
UP & Turn & DD \& PC & 369 & Und & 0.2 \\
Dul & Turn & DD \& PC & 2811 & Int/Spe & 0.2 \\ \midrule
OS & Dial & FED-Dial & 200 & Coh & 0.2  \\
UR & Dial & FED-Dial & 200 & Coh & 0.1 \\ \midrule
SC & Dial & DECODE & 200 & Eng & 0.1 \\
UD & Dial & FED-Dial & 200 & Eng & 0.1 \\
RO & Dial & FED-Dial & 200 & Eng & 0.1 \\\midrule
GU & Dial & FED-Dial & 200 & Eng/Inf & 0.2  \\
CR & Dial & FED-Dial & 200  & Inf & 0.2 \\
\bottomrule
\end{tabular}}
\caption{Adversarial Perturbation Data Statistics. DD \& PC refer to DailyDialog \& PersonaChat Respectively. RR, RP, RNE, Con, Rep, UP, Dul, OS, UR, SC, UD, RO, GU, and CR refer to Random Response, Replace Pronoun, Replace Named Entity, Contradiction, Repetition, Unnatural Paraphrase, Dullness, Order Shuffle, Utterance Replacement, Self-Contradiction, Utterance Duplication, Repeating Others, Generic Utterance, and Content Reduction respectively.}
\label{tab:perturbation-statistics}
\end{table}

\begin{table*}[!ht]	
\centering
\small
\resizebox{0.85\linewidth}{!}{
\begin{tabular}{@{}l|ccccccc@{}}
\toprule
\multicolumn{8}{c}{Turn Level} \\ \midrule
\textbf{Perturbations}  & \textbf{Palm-2} & \textbf{ChatGPT} & \textbf{Tulu} & \textbf{Chimera} & \textbf{Baize} &  \textbf{WizardLM} & \textbf{L2-Chat} \\ \midrule
Random Response ($\downarrow$Rel) & \textbf{0.623} &0.391 & 0.419 & 0.307 & 0.043 & 0.125 & 0.429 \\
Replace Pronoun ($\downarrow$Rel)& \textbf{0.388} & 0.189 & 0.156 & 0.249&0.041 & 0.083 & 0.272\\
Replace Named Entity ($\downarrow$Rel)& \textbf{0.391} &0.160& 0.274&0.301 &0.097 & 0.156& 0.337\\
Contradiction ($\downarrow$Rel)&0.529 & 0.102& 0.523& \textbf{0.535}&0.214 & 0.282 & 0.529 \\
Repetition ($\downarrow$Und)&0.127& \textbf{0.139} &0.060& 0.068& 0.003& 0.002 & 0.016\\
Unnatural Paraphrase ($\downarrow$Und)& 0.307& \textbf{0.451} & 0.182&0.252&0.014 & 0.008& 0.065\\ 
Dullness ($\downarrow$Int)&\textbf{0.433} &0.107 &0.260 & 0.122& 0.000& 0.128 & 0.245 \\ 
Dullness ($\downarrow$Spe)&\textbf{0.411} &0.264&0.191 & 0.092& 0.000& 0.023 &0.166 \\
\midrule
Average & \textbf{0.401} &0.225 &0.258 & 0.241&0.052&  0.101&  0.257 \\ \midrule
\multicolumn{8}{c}{Dialogue Level} \\ \midrule
Order Shuffle ($\downarrow$Coh)&0.440&\textbf{0.535}&0.135&0.000&0.000&0.000&0.155\\
Utterance Replacement ($\downarrow$Coh)&0.490&\textbf{0.730}&0.130&0.000&0.000&0.000&0.075 \\
Generic Utterance ($\downarrow$Eng) &0.295&0.145&0.110&0.085&0.000&0.050&\textbf{0.620}  \\ 
Self-Contradiction ($\downarrow$Eng)&0.365&0.075&\textbf{0.645}&0.455&0.050&0.155&0.535\\ 
Utterance Duplication ($\downarrow$Eng)&0.230&0.095&0.235&0.150&0.000&0.030&\textbf{0.665}\\ 
Repeating Others ($\downarrow$Eng)&0.285&0.090&0.120&0.255&0.025&0.060&\textbf{0.445} \\ 
Content Reduction ($\downarrow$Inf)&\textbf{0.355}&0.040&0.235&0.170&0.000&0.055&0.230 \\
Generic Utterance ($\downarrow$Inf)&0.330&0.065&0.210&0.155&0.000&0.050&\textbf{0.365}\\ 
\midrule
Average &0.349&0.222&0.228&0.159&0.009&0.050&\textbf{0.386} \\
\bottomrule
\end{tabular}}
\caption{Percentage of cases when the LLMs successfully detect a perturbation (Robustness Ratio $R$). The best ratio for each perturbation is highlighted in bold. All the open-source LLMs are the 13B variants and L2-Chat refers to LLaMA-2-Chat-13B.}
\label{tab:adversarial-results}
\end{table*}


\section{Robustness of the LLM Evaluators}
\label{sec:metric-robustness}

Motivated by prior works on applying perturbation strategies for metric robustness probing~\citep{sai-etal-2021-perturbation,khalid-lee-2022-explaining}, we analyze the robustness of LLM evaluators with a series of adversarial strategies and table~\ref{tab:perturbation-statistics} presents the data sources and statistics of our perturbation test suit\footnote{The detailed definitions of each perturbation strategy are outlined in the Appendix.}. 

We focus on negative adversarial perturbations that diminish the quality of the original response or dialogue. Formally, let \( q_{dim} \) represent the score assigned by LLMs for a high-quality dialogue/response specific to a certain dimension. Conversely, \( p_{dim} \) denotes the score given by the LLMs when a particular negative perturbation, targeting that dimension, is applied to the response or dialogue. The LLMs' robustness against that particular perturbation can be determined by $\text{R} =\frac{1}{N}\sum{y}$. $N$ is the number of data instances generated with that particular perturbation and $y$ is calculated as 

\[
      y =
    \begin{cases}
      1 & \text{if $q_{dim} - p_{dim} > \theta$}\\
      0 & \text{otherwise}
    \end{cases}  
\]
where \( \theta \) is a positive threshold value determining the extent of quality reduction introduced by the perturbation. It's worth noting that some perturbation strategies result in greater quality degradation than others. A larger $R$ signifies greater robustness. As that robustness analysis is only meaningful when applied to strong automatic metrics, we limit the analysis to ChatGPT, Palm-2 Bison, Tulu-13B, Chimera-inst-chat-13B, Baize-v2-13B, WizardLM-13B-V1.2, and LLaMA-2-Chat-13B.

\subsubsection{Impact of Different Strategies} 

For the "random response" perturbation at the turn level, Palm-2 Bison, the top-performing LLM, scores the original response over 0.3 higher than the perturbed response in 62.3\% of cases. The other LLMs manage this less than half the time. Notably, Baize-v2-13B and WizardLM-13B-V1.2 frequently fail to recognize the "random response" perturbation. The ``replace pronoun" and "replace named entity" perturbations are more challenging than "random response" as the robustness ratio, $R$, of all the LLMs drops to below 40\%. These two strategies require a more fine-grained understanding of the semantics of the dialogue context and the response. For the ``contradiction" perturbation, Palm-2 Bison and three other open-source LLMs, Tulu-13B, Chimera-13B, and LLaMA-2-Chat-13B achieve a robustness ratio of more than 50\%, significantly outperforming ChatGPT.

In general, all the LLMs perform poorly on ``repetition", ``unnatural paraphrase", and ``dullness" perturbations. The observation is in line with their weaker correlations in the interestingness and understandability dimensions than in the relevance dimension as shown in Table~\ref{tab:full-corr-results}. Notably, ChatGPT performs much better than other LLMs in handling the  ``unnatural paraphrase" perturbation.

The proprietary models are more adept at handling the perturbations targeting dialogue-level coherence than the open-source ones. For example, ChatGPT achieves a robustness ratio of 0.535 and 0.730 for ``Order Shuffle" and ``Utterance Replacement" respectively. Most LLMs struggle with perturbations targeting dialogue-level engagingness and informativeness, such as ``Utterance Duplication",  ``Repeating Others", and ``Generic Utterance", except for LLaMA-2-Chat-13B, suggesting that future research on LLMs for automatic dialogue evaluation should prioritize a deeper comprehension of multi-turn dialogues, such as the depth of topic engagement, speaker sentiments, interactivity \& proactivity among the speakers, etc., which goes beyond mere surface-level coherence assessments.

\subsubsection{Proprietary vs Open-Source LLMs} 

As illustrated in Table~\ref{tab:adversarial-results}, Palm-2 Bison exhibits superior robustness at the turn level, whereas LLaMA-2-Chat-13B performs the best at the dialogue level. Palm-2 Bison can handle most of the adversarial perturbations. Notably, it excels in identifying declines in response relevance, interestingness, specificity, and dialogue coherence. ChatGPT is capable of dealing with adversarial perturbations targeting response understandability and dialogue coherence. Surprisingly, it performs poorly in handling other types of perturbations. Among the open-source LLMs, Tulu-13B and LLaMA-2-Chat-13B perform similarly on average at the turn level. They are better than the other three open-source models. At the dialogue level, LLaMA-2-Chat-13B performs exceptionally well and outperforms Palm-2 Bison by 3.7\% and ChatGPT by 16.4\% on average. It demonstrates consistent strength in dealing with all the perturbations except those targeting response understandability and dialogue coherence. In contrast, both Baize-13B and WizardLM-13B struggle to handle negative perturbations at both the turn and dialogue levels. 

In general, we can observe that none of the LLMs are robust against all the adversarial perturbations. The variance in performances underscores the inherent complexity of dialogues and the challenges in creating a universally robust automatic dialogue evaluator. Future work should prioritize building on these findings to improve the robustness and adaptability of LLMs across diverse perturbations.

\section{Related Work}

\citet{huynh2023understanding} conduct a comprehensive analysis of the dialogue evaluation capability of LLMs with varying model types, sizes, choices of training data, etc. Their study is limited to correlation analysis on a few LLMs, which are mainly vanilla models without instruction-tuning, such as BLOOM~\citep{workshop2023bloom} and OPT~\citep{zhang2022opt}. With the increasing popularity of API-based proprietary instruction-following LLMs, such as OpenAI's ChatGPT and Anthropic Claude~\citep{bai2022constitutional}. Several recent works~\cite{chen2023exploring,liu2023gpteval,lin-chen-2023-llm} study the dialogue evaluation capability of these LLMs via prompting and show that such LLMs exhibit strong zero-shot correlations with human evaluation. Yet, their study is constrained in terms of the number of meta-evaluation datasets, mode of assessment, and number of LLMs examined. Our research addresses these constraints, providing profound insights into the use of various LLMs for automatic dialogue evaluation. We conduct an extensive multi-dimensional correlation analysis involving 30 of the latest popular LLMs and introduce a range of perturbation strategies to assess their robustness.

\section{Conclusion}

In summary, we've analyzed the multi-dimensional evaluation abilities of 30 recent LLMs, covering coherence, engagingness, and more at dialogue and turn levels.  To facilitate the analysis, we enrich 12 existing meta-evaluation datasets with new annotations, which will be publicly available for benchmarking new metrics. Beyond correlation analysis, we introduce various adversarial strategies to test LLM robustness, a perspective not explored in existing works. Lastly, we also examine the impact of dimension-wise and model-wise ensembles on dialogue evaluation in our work. The key insights are summarized as follows:
    \begin{enumerate}
        \item Instruction-tuned models align better with human evaluations than vanilla foundation models.
        \item Proprietary models, especially GPT-4,  have superior evaluation abilities compared to open-source LLMs.
        \item Model size and instruction data are vital for evaluation. Only the ensemble of strong open-source models performs on par with ChatGPT and PaLM-2 Bison.        
        \item LLMs excel more in evaluating coherence, relevance, and overall quality than specificity and diversity. Using an ensemble of their dimension-specific scores aligns better with the overall human evaluations than a direct assessment of the overall quality.
        
        \item None of the LLMs are robust against all the adversarial perturbations. Google's Palm-2 Bison achieves the best robustness at the turn level while Meta's LLaMA-2-Chat-13B~\citep{touvron2023llama2} tops at the dialogue level. 
        \item  LLMs show promise in automatic dialogue evaluation, but it's still an open problem, with even GPT-4 not excelling in all dimensions (achieve correlations $> 0.8$).
    \end{enumerate}

\section{Acknowledgments}

We thank the anonymous reviewers for their insightful comments. This work is supported by Human Robot Collaborative AI under its AME Programmatic Funding Scheme (Project No. A18A2b0046), the National Natural Science Foundation of China (Grant No. 62271432, 62106038), Shenzhen Science and Technology Research Fund (Fundamental Research Key Project Grant No. JCYJ20220818103001002), and the Internal Project Fund from Shenzhen Research Institute of Big Data under Grant No. T00120220002. This work is also a result of the projects: ASTOUND (101071191 - HORIZON-EIC-2021-PATHFINDERCHALLENGES-01) funded by the European Commission, BEWORD (PID2021-126061OB-C43) funded by MCIN/AEI/10.13039/501100011033 and, as appropriate, by “ERDF A way of making Europe”, by the “European Union”, and the Research Grants for Young Investigators from Universidad Politécnica de Madrid (GENIUS:APOYO-JOVENES-21-TAXTYC-32-K61X37) funded by Comunidad de Madrid.

\appendix

\section{Additional Details on Meta-Evaluation Datasets}
\label{sec:additional-details-datasets}

\subsection{Dialogue-Level Datasets}

Due to reasons such as containing only single-annotator ratings and ratings of poor inter-annotator agreements, we perform re-annotation of all the dialogue-level datasets except for FED-Dial~\citep{mehri-eskenazi-2020-unsupervised} with GPT-4. Due to the large number of dialogues in IEval-Dial~\citep{svikhnushina-etal-2022-ieval}, Persona-See~\citep{see-etal-2019-makes}, and Reliable-Dial~\citep{ji-etal-2022-achieving} and the high API costs of GPT-4, we randomly sample 500 dialogues from each for the re-annotation.

\subsubsection{IEval-Dial~\citep{svikhnushina-etal-2022-ieval}} is constructed to benchmark state-of-the-art empathetic chatbots, such as BlenderBot~\citep{roller-etal-2021-recipes} and MIME~\citep{majumder-etal-2020-mime}. All dialogues are grounded on an emotional situation. The experiencer who interacts with the chatbot will rate the multi-turn conversation and rank the multi-turn interactions from four different chatbots simultaneously. The overall rating of a conversation is based on the ordinal ranking of the chatbot among all the chatbots grounding on a particular emotional situation and emotion polarity (positive vs negative). In total, IEval-Dial contains 1920 multi-turn human-chatbot conversations.

\subsubsection{Persona-See~\citep{see-etal-2019-makes}} contains the most number of human-chatbot multi-turn dialogues among all the English evaluation datasets (3316 dialogues). It covers 29 dialogue models, which are trained on the PersonaChat~\citep{zhang-etal-2018-personalizing} with various controlled settings, including repetition, response-relatedness, specificity, and question-asking. 

\subsubsection{Reliable-Dial~\citep{ji-etal-2022-achieving}} contains three data splits with 2925 multi-turn human-chatbot dialogues in total. The data are generated by crowdsource workers interacting with 10 different dialogue models that are pretrained with the PersonaChat dataset~\citep{zhang-etal-2018-personalizing}, such as Poly-Encoder Transformer and Key-Value Memory Network. The three data splits are collected in three rounds of human evaluation replication experiments: ice-breaker, free run 1, and free run 2. In the ice-breaker round, the topic for discussion was directly selected from the persona of the dialogue model. In the free run, the crowdsource workers are free to input the topics of the conversations.

\subsubsection{HEval-Dial~\citep{smith-etal-2022-human}}~\citet{smith-etal-2022-human} proposes five different human evaluation settings, including per-turn pairwise comparison, per-turn single model rating, dialogue-level pairwise comparison, self-chat dialogue-level pairwise comparison, and single model dialogue-level rating. The HEval-Dial dataset is collected under the last setting and it consists of 286 multi-turn human-chatbot conversations. The dataset covers four different dialogue models, including BlenderBot3B, BlenderBot3B-M0 (BlenderBot3B without output length control), BlenderBot90M (smaller model than BlenderBot3B), and Reddit3B (without finetuning on high-quality dialogue data).

\subsubsection{FED-Dial~\citep{mehri-eskenazi-2020-unsupervised}} consists of 125 dialogues. Among the dialogues, 40 are collected between a human and the Meena chatbot, 44 are collected between a human and the Mitsuku chatbot, and the rest are human-human dialogues. Each dialogue in FED is annotated by five human judges across 11 quality dimensions: coherence, error recovery, consistency, diversity, topic depth, likability, understanding, flexibility, informativeness, inquisitiveness, and overall impression. The ratings of all the dimensions are based on the 1-3 Likert scale except that the consistency and overall ratings range from 0 to 1 and 1 to 5 respectively. The inter-annotator agreements for all dimensions are above 0.8 in terms of Spearman correlations except consistency (0.562), diversity (0.789), and inquisitiveness (0.769).

\subsubsection{ConTurE-Dial~\citep{mehri-etal-2022-interactive}} is a subset of the DSTC9 Interactive benchmark~\citep{mehri-etal-2022-interactive}, which measures how well a dialogue system can hold a cohesive multi-turn conversation across 11 quality dimensions (same as those in FED-Dial). The dataset is collected for the "Interactive Evaluation of Dialog" task of the DSTC9 challenge\footnote{The Ninth Dialogue System Technology Challenge}. There are 2200 human-chatbot dialogues from 10 dialogue models with each annotated by 3 annotators.~\citet{ghazarian-etal-2022-wrong} reuse the 119 dialogues containing less than 10 rounds of human-chatbot exchanges in DSTC9 to form ConTurE-Dial. 

\subsection{Turn-Level Datasets}

\subsubsection{ConTurE-Turn~\citep{ghazarian-etal-2022-wrong}} is curated by~\citet{ghazarian-etal-2022-wrong} and it consists of 119 dialogues from the DSTC9 Interactive benchmark. The authors annotate each chatbot turn for the overall quality. To facilitate multi-dimensional evaluation, we provide fine-grained annotations for response relevance, specificity, understandability, and interestingness with GPT-4.

\subsubsection{Persona-USR \& Topical-USR~\citep{mehri-eskenazi-2020-usr}} are two high-quality human evaluation turn-level datasets with multi-dimensional quality annotations. Persona-USR and Topical-USR contain 300 and 360 context-response pairs and each context-response pair is annotated by three dialogue researchers along six different dialogue quality categories: Understandable (0-1), Natural (1-3), Maintains Context (1-3), Interesting (1-3), Uses Knowledge (0-1), Overall Quality (1-5)\footnote{The numbers in the bracket are the Likert scales.}. The inter-annotator agreements for quality dimensions of both datasets are generally good ($>$ 0.5). We only need to annotate the specificity dimension with GPT-4.

\subsubsection{DailyDialog-Zhao~\citep{zhao-etal-2020-robust}} evaluation dataset is collected based on 100 dialogue contexts from the test set of the DailyDialog corpus~\citep{li-etal-2017-dailydialog}. Conditioned on each context, 9 diverse dialogue models, such as LSTM Seq2SeqAttn and GPT-2, are employed to generate the responses. Four dimensions are annotated: appropriateness, language usage, relevance, and content. Each context-response pair is rated by four annotators on a 5-point Likert scale. The Krippendorff's $\alpha$ along appropriateness after the removal of outliers is above 0.8. We provide additional annotations for interestingness, understandability, and specificity with GPT-4.

\subsubsection{Persona-Zhao~\citep{zhao-etal-2020-robust}}  evaluation dataset is constructed in a similar manner as DailyDialog-Zhao. The context-response pairs of Persona-Zhao are collected based on dialogues from the test set of the PersonaChat corpus~\citep{zhang-etal-2018-personalizing}. Only the appropriateness quality of the response is annotated in Persona-Zhao, with an inter-annotator agreement above 0.8 in terms of Krippendorff's $\alpha$. Hence, we need to annotate the responses along interestingness, specificity, relevance, and understandability.

\subsubsection{FED-Turn~\citep{mehri-eskenazi-2020-unsupervised}} is by far the most comprehensive dataset for multi-dimensional turn-level evaluation. It contains 375 context-response pairs derived from human-chatbot conversations in FED-Dial. Each response is annotated by five crowd-workers for 9 different quality dimensions, which cover all the five dimensions used in our paper. The inter-annotator agreements, in terms of Spearman correlations, are high ($>$ 0.7) for all dimensions, except understandability (0.522). 

\section{Essential Details of LLMs}
\label{sec:llm-details}

Table~\ref{tab:question-design} depicts the detailed question design that we use to prompt open-source LLMs for evaluating different dialogue quality dimensions. Table~\ref{tab:llm-details} provides a summary of the essential details of different LLMs, including their model architecture, training data, etc.

\begin{table}[ht] 
  \centering 
\resizebox{\linewidth}{!}{
    \renewcommand\tabcolsep{2.4pt}
    \begin{tabular}{ll}
    \toprule
    \textbf{Dimension-Specific} & \textbf{Question Design} \\
    \midrule
    \multicolumn{2}{l}{\textbf{\textit{Turn Level}}} \\
    \midrule
    \multicolumn{1}{l}{\multirow{1}[1]{*}{Interestingness}}  & Question: Is the responses interesting? \\
    \cmidrule(lr){2-2}
    \multicolumn{1}{l}{\multirow{1}[1]{*}{Understandability}}   & Question: Is the responses understandable? \\
    \cmidrule(lr){2-2}
    Relevance & Question: Is the responses relevant to the context? \\
    \cmidrule(lr){2-2}
    Specificity & Question: Is the responses specific to the context? \\
    \cmidrule(lr){2-2}
    Overall & Question: Is the overall quality of the response good? \\
    \midrule
    \multicolumn{2}{l}{\textbf{\textit{Dialogue Level}}} \\
    \midrule
        \multicolumn{1}{l}{\multirow{1}[1]{*}{Coherence}}  & Question: Is the overall flow of the conversation coherent? \\
    \cmidrule(lr){2-2}
    \multicolumn{1}{l}{\multirow{1}[1]{*}{Engagingness}}   & Question: Is the conversation engaging? \\
    \cmidrule(lr){2-2}
    Informativeness & Question: Are the responses of the chatbot informative? \\
    \cmidrule(lr){2-2}
    Diversity & Question: Are the responses of the chatbot diverse? \\
    \cmidrule(lr){2-2}
    Overall & Question: Is the overall quality of the entire conversation good? \\
    \bottomrule
    \end{tabular}
}
    \caption{Dimension-specific question design.}
  \label{tab:question-design}
\end{table}

\begin{table*}[!ht]	
\centering
\small
\resizebox{\linewidth}{!}{
\begin{tabular}{@{}l|ccccccc@{}}
\toprule
\textbf{Name} & \textbf{Backbone} & \textbf{Architecture} &\textbf{Training Data} & \textbf{Variants} & \textbf{Context Length} &\textbf{SFT} & \textbf{RLHF} \\ \midrule
LLaMA~\citep{touvron2023llama} & - & Decoder-Only & \thead{English CommonCrawl, Wikipedia \\ Github, Gutenberg and Books3, \\ C4, ArXiv, Stack Exchange} & 7B, 13B, 30B, 65B & 2048 & No & No \\ \midrule
LLaMA-2~\citep{touvron2023llama2}& - & Decoder-Only & Unknown & 7B, 13B, 70B & 4096 & No & No\\ \midrule
XGen~\citep{XGen}& - & Decoder-Only & \thead{RedPajama, C4, Wikipedia, \\ Pile\_DM\_Mathematics, Apex code} & 7B & 4K, 8K & No & No\\ \midrule
Falcon~\citep{falcon40b}& - & Decoder-Only & \thead{RefinedWeb, Books, \\ Reddit, arXiv, etc}  & 7B, 40B & 2048 & No & No \\ \midrule
MPT & - & Decoder-Only & \thead{C4, RedPajama, \\ The Stack, S2ORC} & 7B, 30B & 4096 & No & No\\ \midrule
OpenLLaMA~\citep{openlm2023openllama} & - & Decoder-Only & \thead{RedPajama, Falcon Refined-web, \\ StarCoder} & 3B, 7B, 13B & 2048 & No & No \\ \midrule
Pythia~\citep{biderman2023pythia} & - & Decoder-Only & The Pile & \thead{70M, 160M, 410M, \\ 1B, 1.4B, 2.8B, \\ 6.9B, 12B} & 2048 & No & No \\ \midrule
BLOOM~\citep{workshop2023bloom} & - & Decoder-Only & the ROOTS corpus & 560M, 1B, 7B, 3B & 2048 & No & No \\
\midrule
Alpaca~\citep{alpaca} & LLaMA & Decoder-Only & \thead{52K instruction-following \\ demonstrations distilled from OpenAI’s \\ text-davinci-003 with self-instruct}  & 7B, 13B & 2048 & Yes & No \\ \midrule
Vicuna~\citep{vicuna2023} & LLaMA & Decoder-Only & 70K ShareGPT data & 7B, 13B & 2048 & Yes & No \\ \midrule
LLaMA-2-Chat~\citep{touvron2023llama2} & LLaMA-2 & Decoder-Only & \thead{Meta's self-collected instruction \\ data and human feedback data \\ optimized for helpfulness \& safety} & 7B, 13B & 4096 & Yes & Yes \\ \midrule

Baize~\citep{xu2023baize} & LLaMA & Decoder-Only & Self-chat data from ChatGPT & 7B, 13B & 2048 & Yes & No \\ \midrule

Falcon-Ins~\citep{falcon40b} & Falcon & Decoder-Only & Baize Data, GPT4All, GPTeacher  & 7B, 40B & 2048 & Yes & No \\ \midrule

Tulu~\citep{wang2023far} & LLaMA & Decoder-Only & \thead{SuperNI, CoT, Flan V2, Alpaca Data, \\ Dolly, Open Assistant, Self-instruct, \\ Unnatural Instructions, Baize, ShareGPT}  & 7B, 13B & 2048 & Yes & No \\ \midrule

Chimera~\citep{chen2023phoenix} & LLaMA & Decoder-Only & \thead{ShareGPT, Self-instruct, \\ User-centered instructions}  & 7B, 13B & 2048 & Yes & No \\ \midrule

Phoenix~\citep{chen2023phoenix} & BLOOM & Decoder-Only & \thead{Multilingual Alpaca data, Multilingual \\ ShareGPT, Self-generated User-centered \\ multi-lingual instructions} & 7B, 13B & 2048 & Yes & No \\ \midrule

Oasst-sft-Pythia-12B & Pythia & Decoder-Only & OpenAssistant OASST1 & 12B & 2048 & Yes & No \\ \midrule

Dolly-v2~\citep{DatabricksBlog2023DollyV2} & Pythia & Decoder-Only & Databricks Dolly-15k & 3B, 7B, 12B & 2048 & Yes & No \\ \midrule

MPT-7B-8K-Instruct & MPT & Decoder-Only & \thead{Databricks Dolly-15, Anthropic \\ Helpful and Harmless (HH-RLHF)} & 7B & 8K & Yes & No \\ \midrule

XGen-8K-7B-Inst & XGen & Decoder-Only & \thead{Databricks Dolly-15, Baize Data, \\ OpenAssistant OASST1, GPT-related datasets)} & 7B & 8K & Yes & No \\ \midrule

ChatGLM-v2~\citep{zeng2023glmb} & GLM & Prefix Decoder & Unknown & 6B & 8K & Yes & No \\ \midrule

WizardLM~\citep{xu2023wizardlm} & LLaMA & Decoder-Only & Evol-Instruct with ChatGPT & 7B, 13B, 30B & 8K & Yes & No \\ \midrule

ChatGPT (gpt-3.5-turbo) & GPT-3 & Decoder-Only & Unknown & Unknown & 4096 & Yes & Yes \\ \midrule

Palm-2 Bison (text-bison-001) & Palm-2 & Decoder-Only & Unknown & Unknown & 8192 & Yes & No \\ \midrule

\end{tabular}}
\caption{Essential details of different LLMs. Variants are presented in terms of the number of parameters (B - billion, M - million). SFT refers to whether the model has been instruction-tuned and RLHF refers to whether the model has been aligned to human feedback data via reinforcement learning.}
\label{tab:llm-details}
\end{table*}

\section{Additional Analysis}
\label{sec:additional-analysis}

\subsubsection{Inter-Dimensional Correlations of Different LLMs On FED-Dial} 
\begin{figure*}[ht!] 
\begin{subfigure}{0.32\textwidth}
\includegraphics[width=\linewidth]{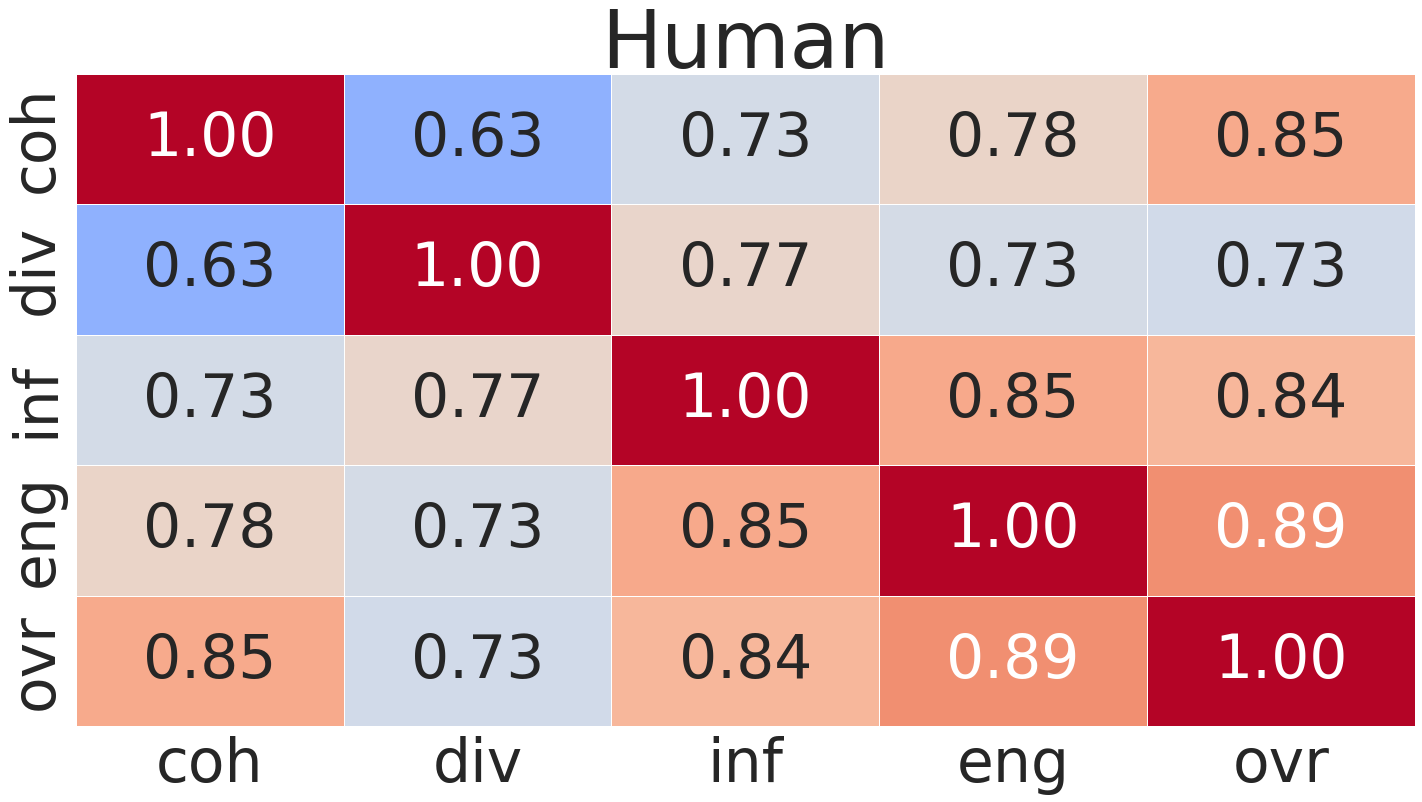}
\label{fig:a}
\end{subfigure}\hspace*{\fill}
\begin{subfigure}{0.32\textwidth}
\includegraphics[width=\linewidth]{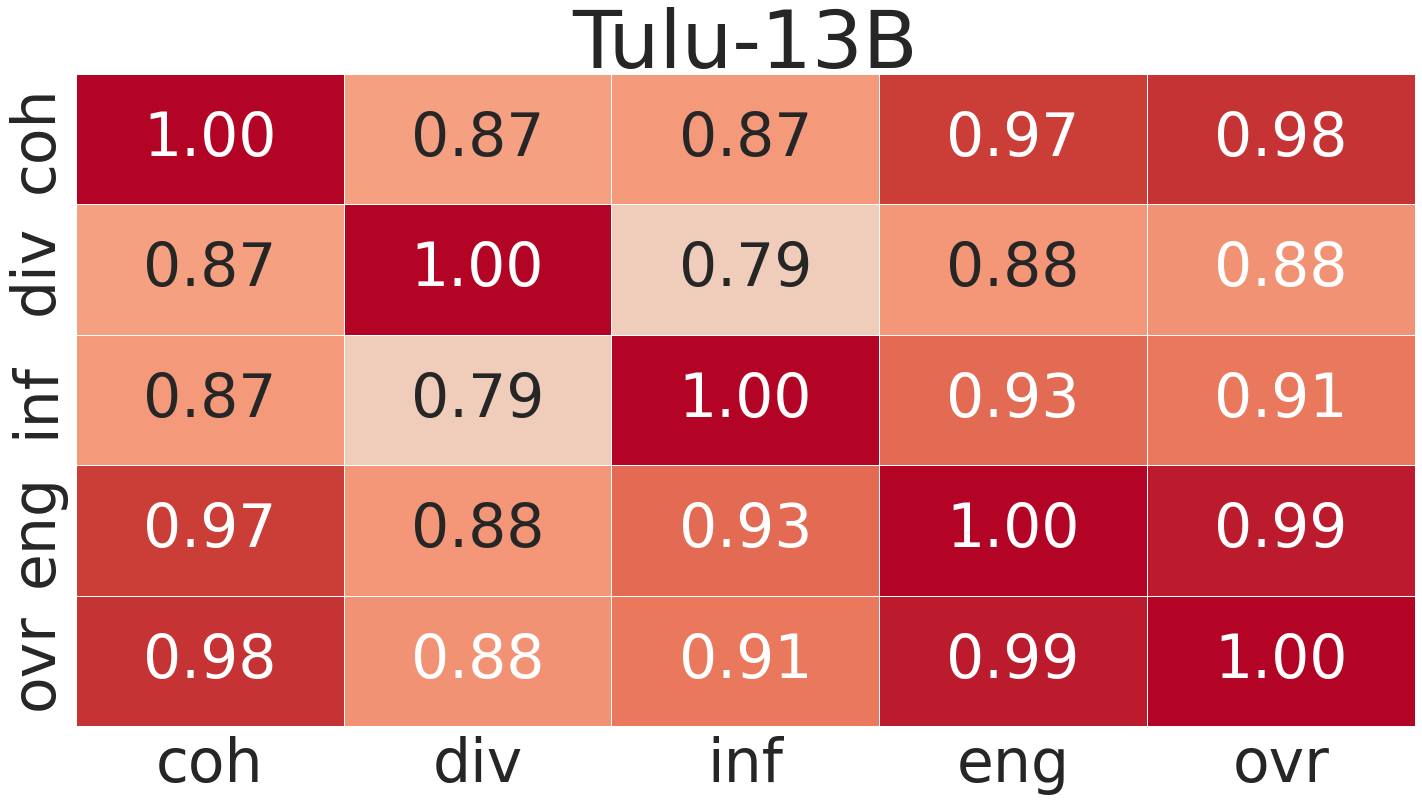}
\label{fig:b}
\end{subfigure}\hspace*{\fill}
\begin{subfigure}{0.32\textwidth}
\includegraphics[width=\linewidth]{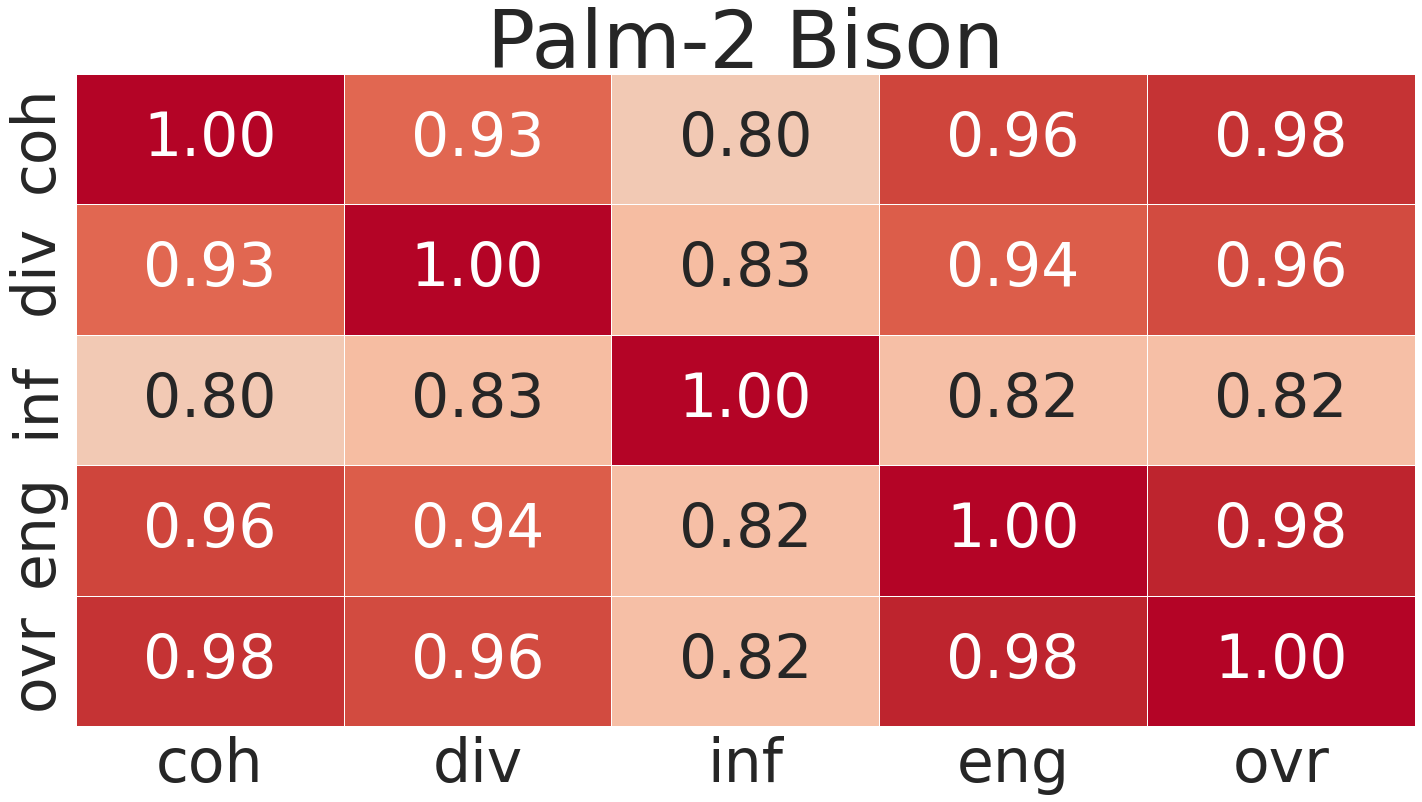}
\label{fig:c}
\end{subfigure}

\medskip
\begin{subfigure}{0.32\textwidth}
\includegraphics[width=\linewidth]{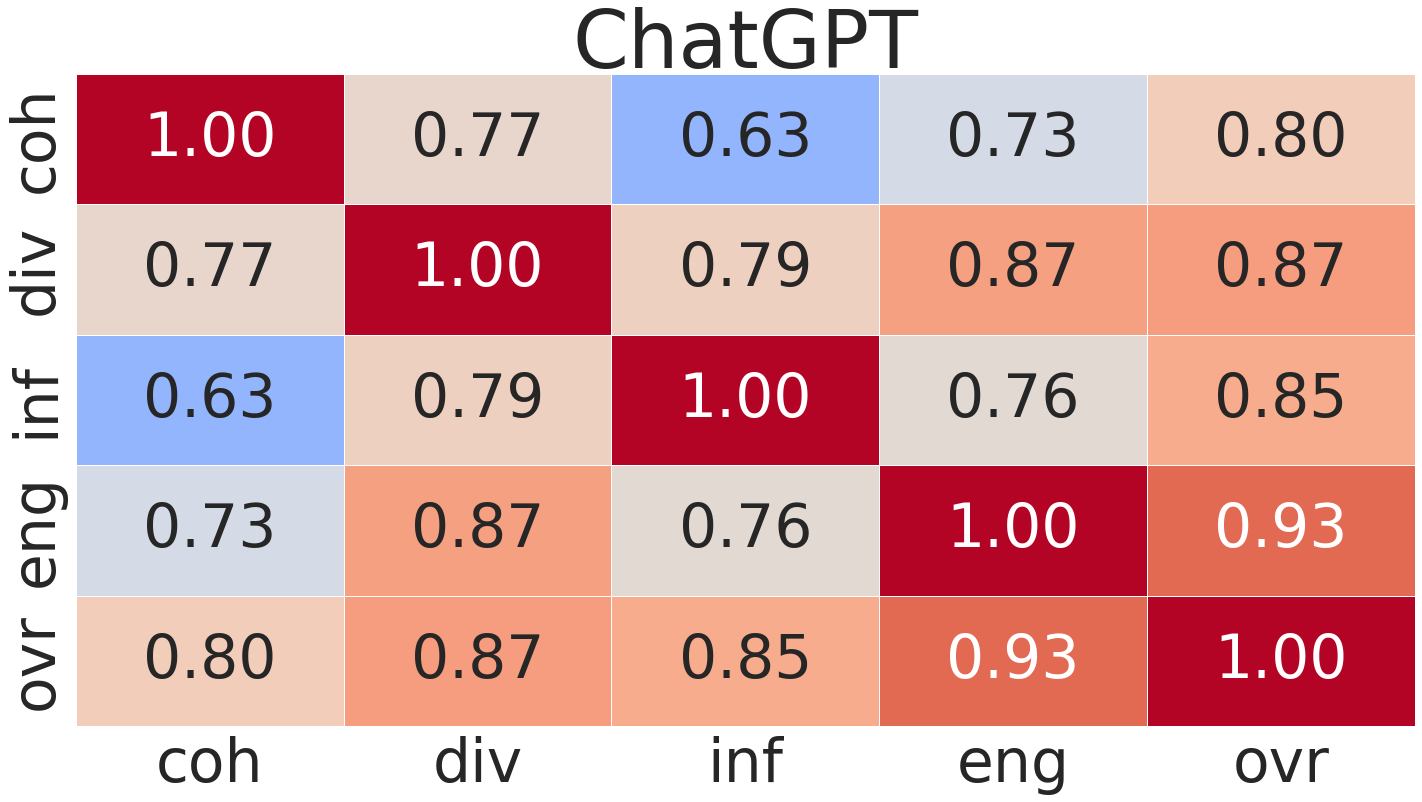}
\label{fig:d}
\end{subfigure}\hspace*{\fill}
\begin{subfigure}{0.32\textwidth}
\includegraphics[width=\linewidth]{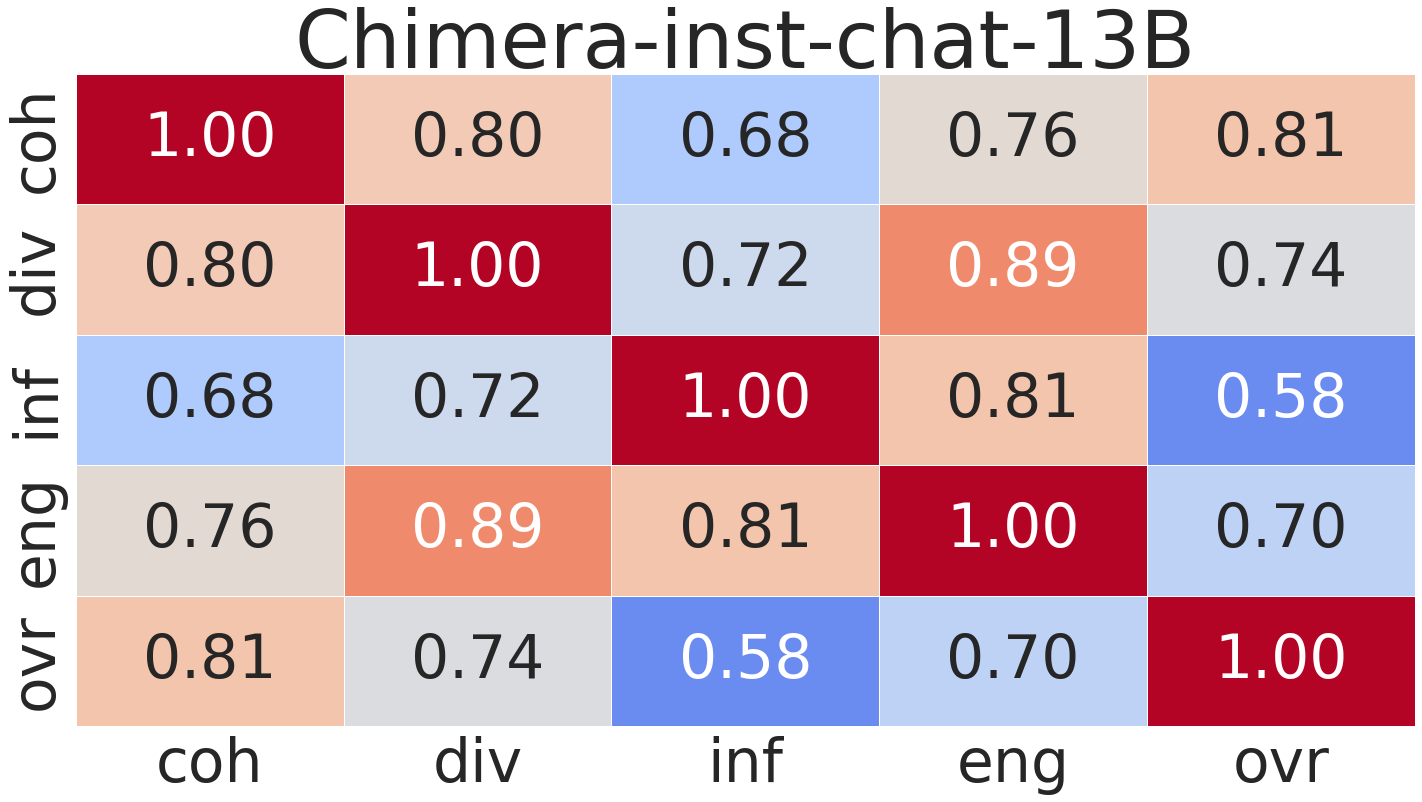}
\label{fig:e}
\end{subfigure}\hspace*{\fill}
\begin{subfigure}{0.32\textwidth}
\includegraphics[width=\linewidth]{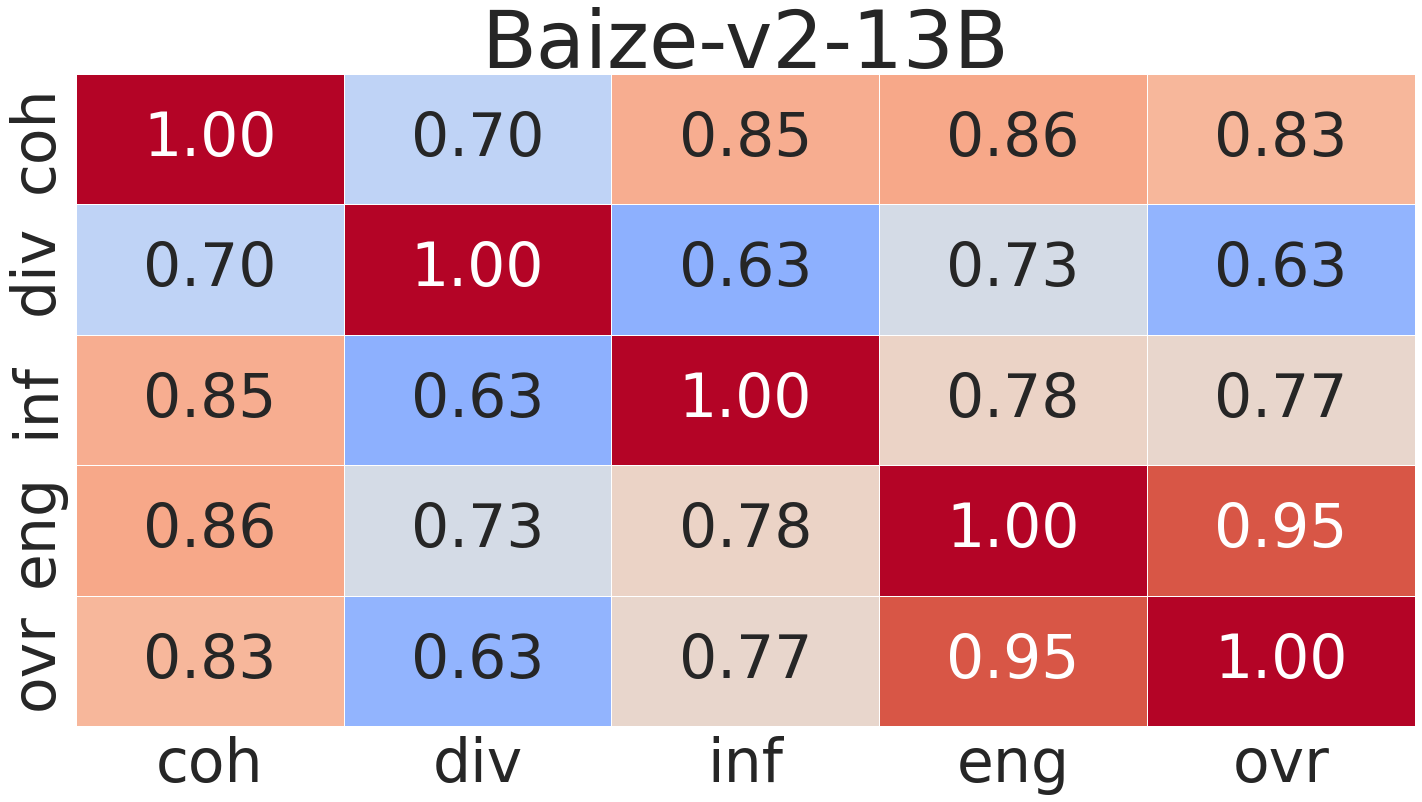}
\label{fig:f}
\end{subfigure}

\caption{Inter-dimensional correlations of the gold human ratings and different model scores on the FED-Dial dataset~\citep{mehri-eskenazi-2020-unsupervised}} \label{fig:1}
\end{figure*}

Figure~\ref{fig:1} presents the inter-dimensional correlation patterns of different LLMs as well as the human judges on the FED-Dial dataset. We can observe that the patterns of Tulu-13B and Palm-2 Bison showcase a high interdependence among different dimension-specific scores. The correlations of some dimension pairs even exceed 0.90. The observations suggest that on the FED-Dial dataset, Palm-2 Bison and Tulu-13B only perform well on a subset of dimensions. We can hypothesize that the ensemble of their dimension-specific scores may not bring significant improvement in their overall judgment. The hypothesis is supported by findings in Table~\ref{tab:dimension-ensemble-results}. 

 The pattern of ChatGPT is the most similar to that of humans. Additionally, ChatGPT achieves the best correlations among almost all the dimensions at the dialogue level as shown in Table~\ref{tab:full-corr-results}. The observations indicate that ChatGPT better captures the nuances in evaluating different fine-grained quality dimensions than other LLMs.

 The metric scores of Chimera and Baize on FED-Dial are more diverse. We can observe a more heterogeneous correlation pattern for both LLMs than for Tulu-13B and Palm-2 Bison. The observation explains why the dimension-level ensemble of Chimera and Baize yields significant improvements in overall evaluation than the direct prompting method as discussed in the dimension-wise ensemble section.

 \subsubsection{Inter-Dimensional Correlations of Different LLMs On FED-Turn} 
\begin{figure*}[ht!] 
\begin{subfigure}{0.32\textwidth}
\includegraphics[width=\linewidth]{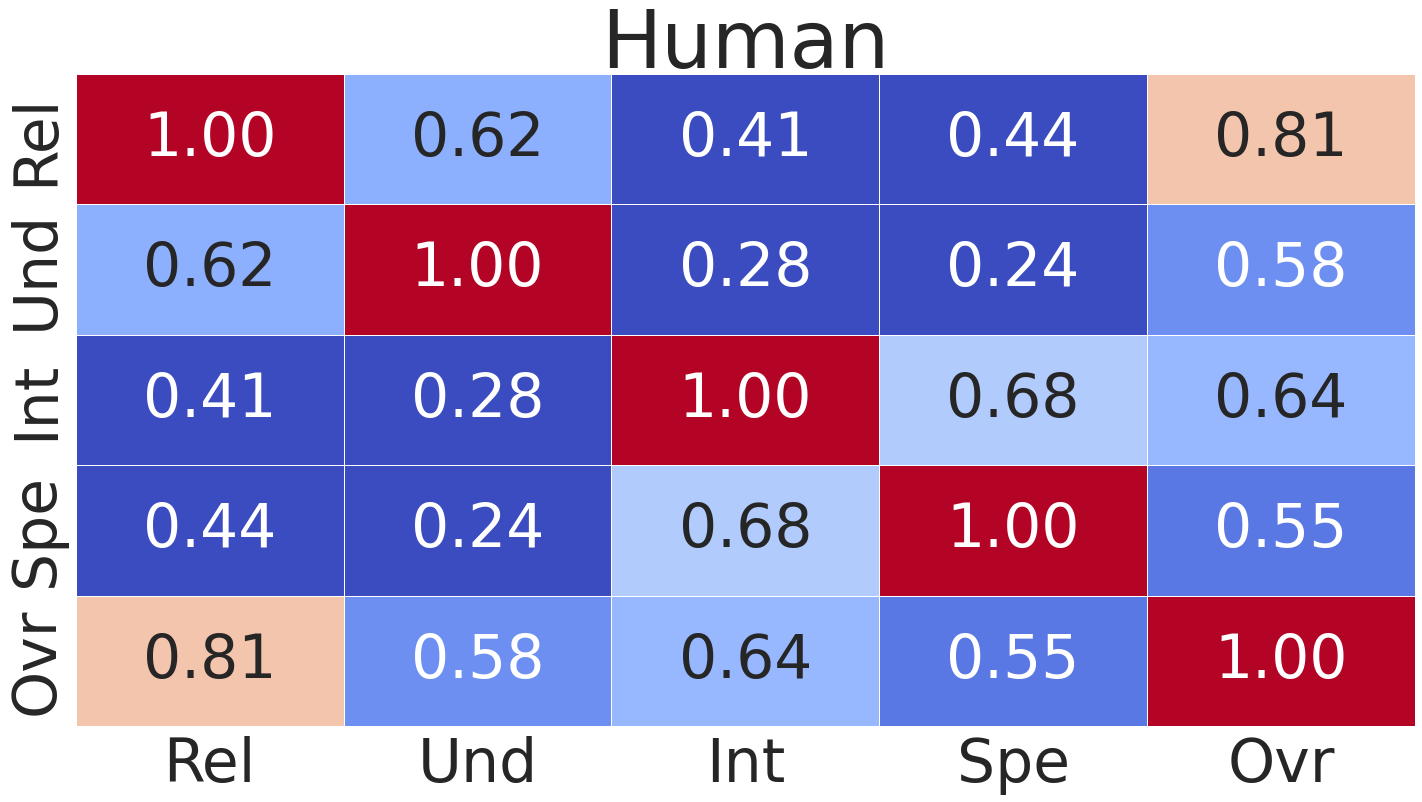}
\label{fig:human_turn}
\end{subfigure}\hspace*{\fill}
\begin{subfigure}{0.32\textwidth}
\includegraphics[width=\linewidth]{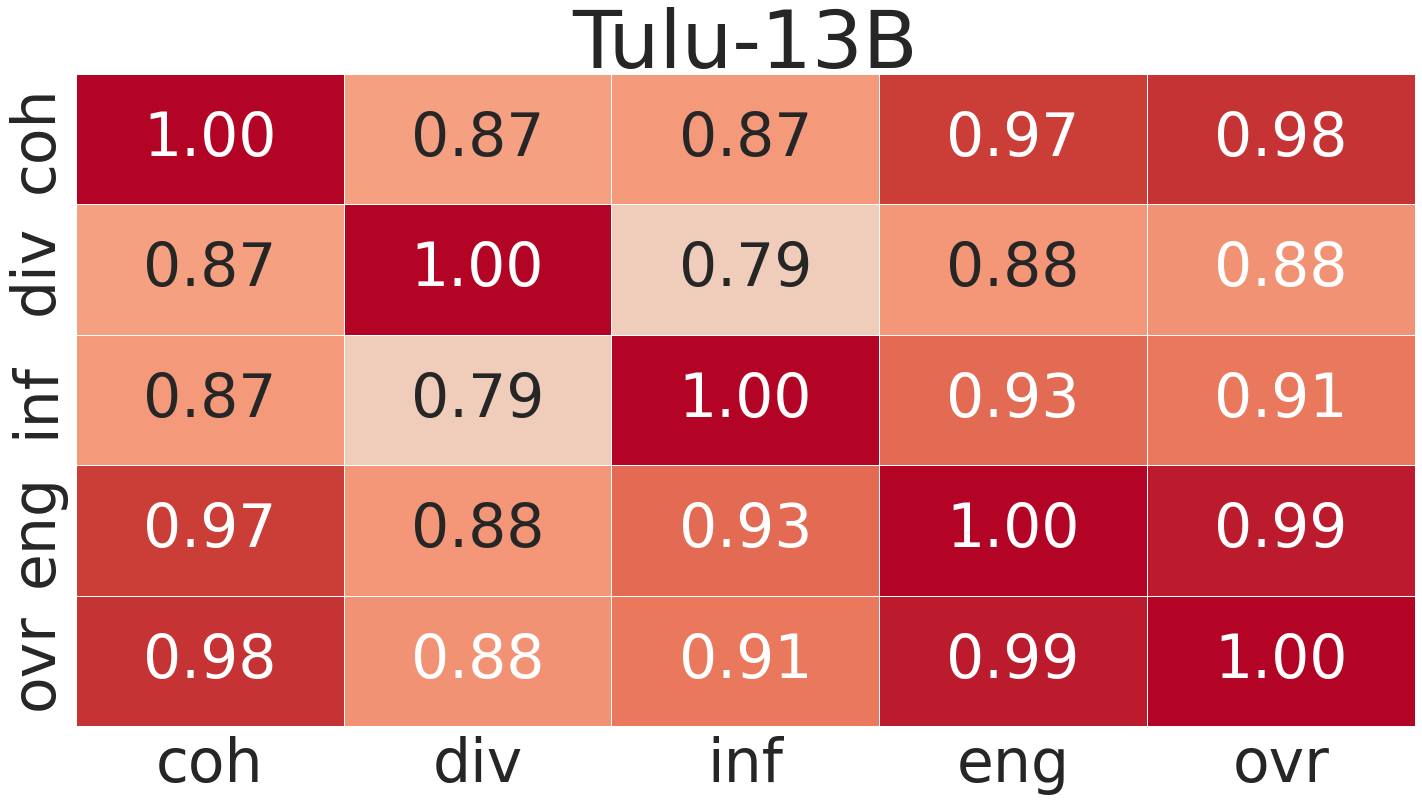}
\label{fig:tulu_turn}
\end{subfigure}\hspace*{\fill}
\begin{subfigure}{0.32\textwidth}
\includegraphics[width=\linewidth]{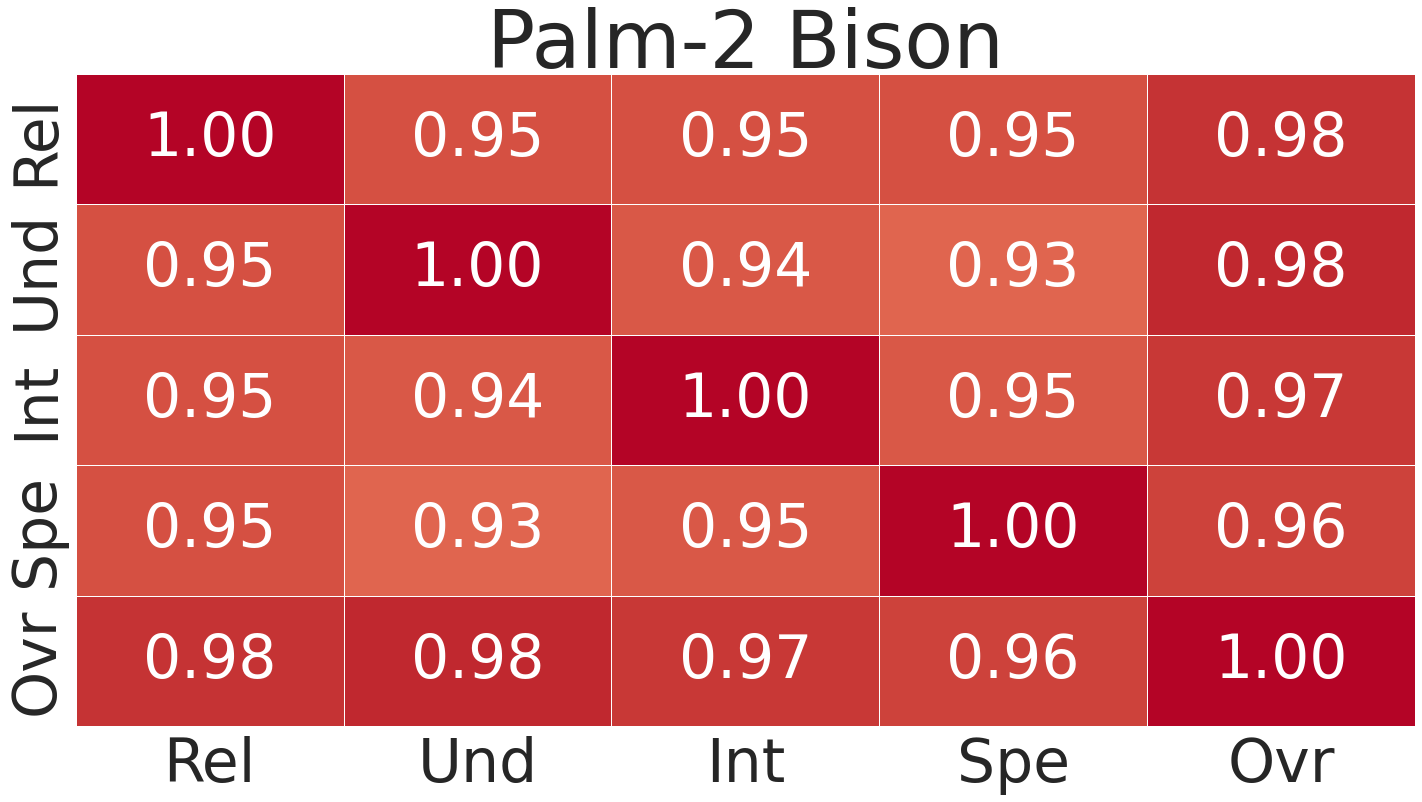}
\label{fig:bison_turn}
\end{subfigure}

\medskip
\begin{subfigure}{0.32\textwidth}
\includegraphics[width=\linewidth]{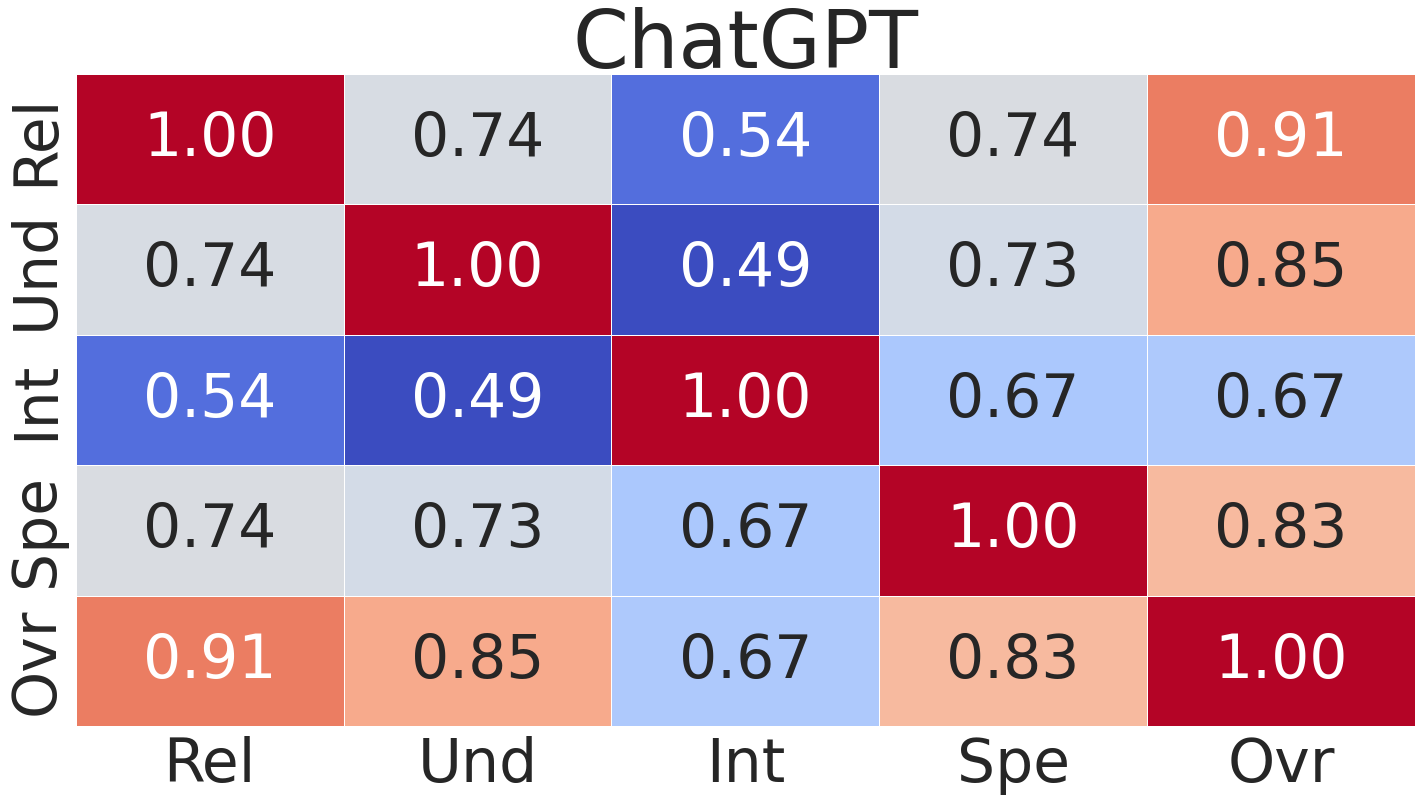}
\label{fig:chatgpt_turn}
\end{subfigure}\hspace*{\fill}
\begin{subfigure}{0.32\textwidth}
\includegraphics[width=\linewidth]{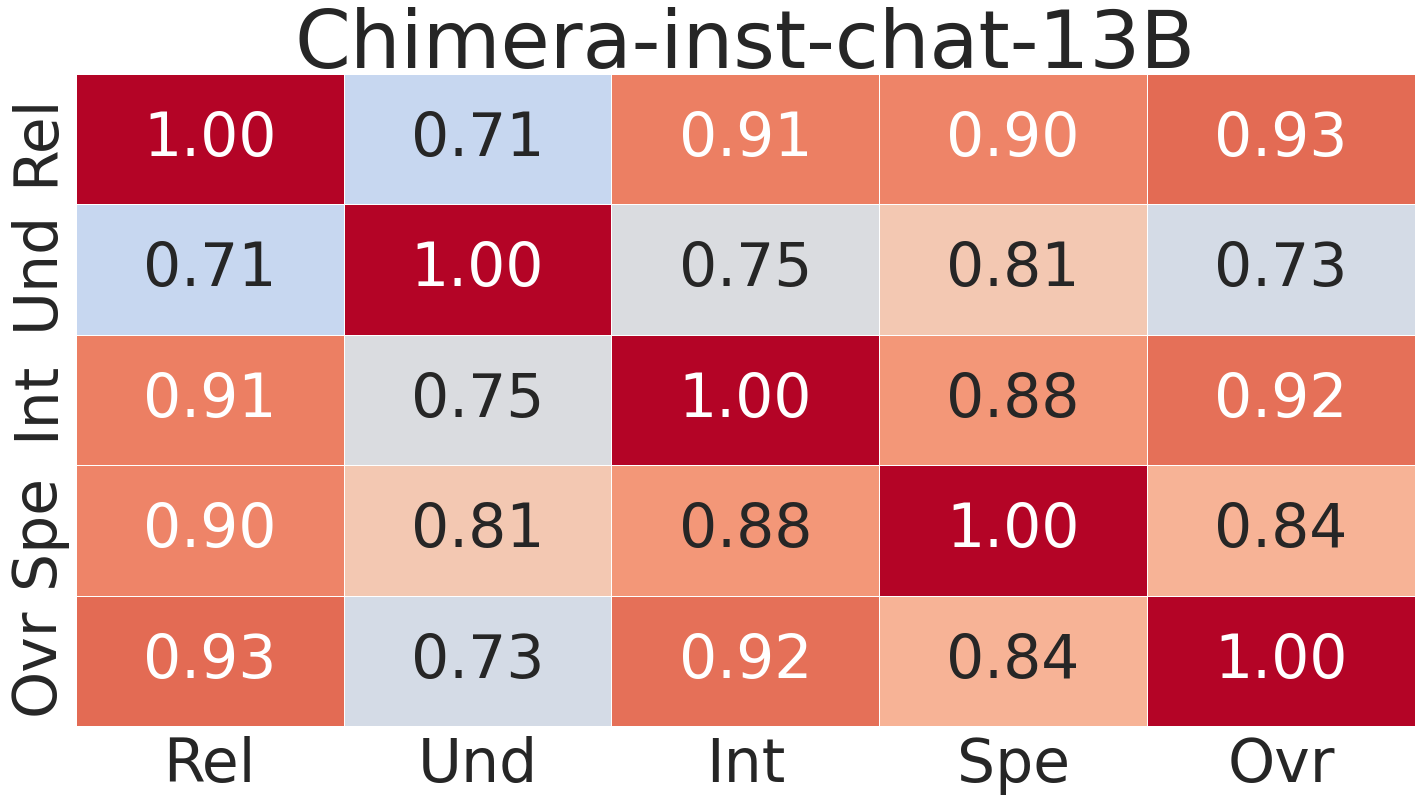}
\label{fig:chimera_turn}
\end{subfigure}\hspace*{\fill}
\begin{subfigure}{0.32\textwidth}
\includegraphics[width=\linewidth]{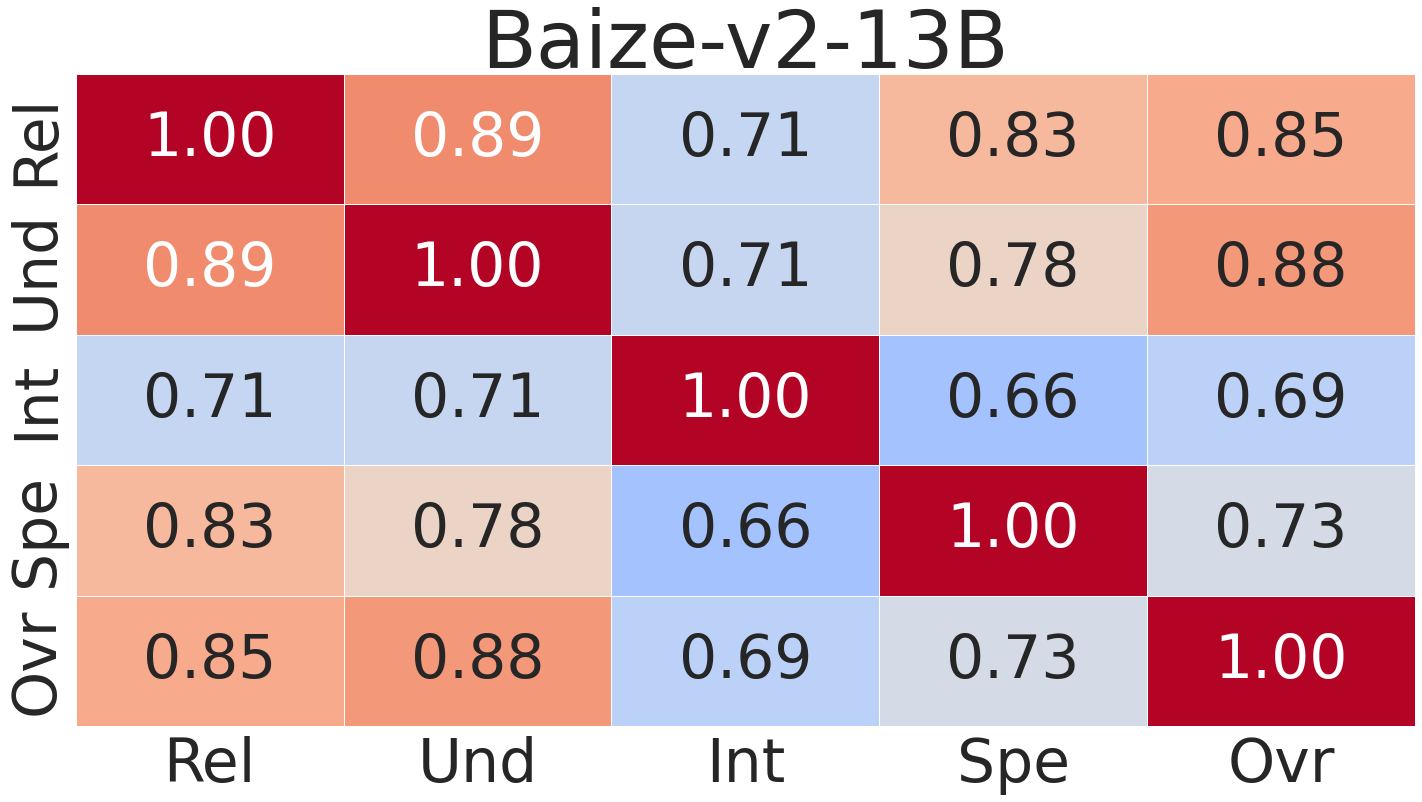}
\label{fig:baize_turn}
\end{subfigure}

\caption{Inter-dimensional correlations of the gold human ratings and different model scores on the FED-Turn dataset~\citep{mehri-eskenazi-2020-unsupervised}} \label{fig:2}
\end{figure*}

Figure~\ref{fig:2} showcases the inter-dimensional correlation patterns of different LLMs as well as the human judges on the FED-Turn dataset. The correlation pattern of ChatGPT on FED-Turn is the most similar to that of humans. The observation explains the best multi-dimensional performance of ChatGPT among all LLMs at the turn level. Regarding Tulu-13B and Palm-2 Bison, we can observe the same patterns on FED-Turn as those displayed on FED-Dial, suggesting that both LLMs are more specialized for evaluating only a subset of the dimensions. Additionally, the metric scores of Chimera and Baize on FED-Turn are more diverse than Tulu-13B and Palm-2 Bison, indicating their better suitability for the dimension-level ensemble at the turn level. As shown in Table~\ref{tab:dimension-ensemble-results}, dimensional-level ensembles bring the most improvements for the overall response evaluation with these two LLMs.

\section{Additional Details of Adversarial Test Data}
\label{sec:adversarial-data-details}

Table~\ref{tab:perturbation-descriptions} provides concise descriptions of the perturbation strategies and the details of each strategy are outlined in the subsequent sections.

\subsection{Turn-Level Adversarial Perturbations}

For the turn-level perturbations, we use the adversarial data\footnote{The data is publicly available at \url{https://github.com/baber-sos/Explaining-Dialogue-Evaluation-Metrics-using-Adversarial-Behavioral-Analysis}.} generated by~\citet{khalid-lee-2022-explaining}. Specifically, we use the splits produced from the human-human conversations in both DailyDialog~\citep{li-etal-2017-dailydialog} and PersonaChat~\citep{zhang-etal-2018-personalizing}. The different turn-level perturbation strategies are listed as follows:

\subsubsection{Relevance Reduction} 

There are four perturbation strategies to reduce the contextual relevance of the responses: 

\smallskip
\noindent\textbf{(a). Random Response} - the response is replaced with a response from other dialogues or the dialogue history. 

\smallskip
\noindent\textbf{(b). Pronoun Replacement} - Techniques for altering pronoun usage: (1) Swap gender-specific pronouns (he/she) with neutral ones (it/they) and vice versa, (2) Change singular pronouns (it/this/I) to plural forms and vice versa, and (3) Replace 1st person pronouns (I/we/our) with random 2nd or 3rd person pronouns and vice versa. 

\smallskip
\noindent\textbf{(c). Named Entities Replacement} - Named entities in a response is modified by (1) substituting them with another random entity of the same category and (2) replacing them with a named entity from a different category. 

\smallskip
\noindent\textbf{(d). Contradiction} - The response's meaning is altered to contradict the dialogue context by (1) negating its verbs, (2) swapping named entities mentioned at least twice in the context with a random entity of the same kind, and (3) replacing co-references of a named entity in the response with a random entity of the same category.

\subsubsection{Understandability Reduction} The following perturbations are adopted to reduce response understandability: 

\smallskip
\noindent\textbf{(a). Repetition} - A random portion of the response is repeated multiple times to make it unnatural. 

\smallskip
\noindent\textbf{(b). Unnatural Paraphrase} - Non-stop English words are randomly sampled and replaced with synonyms from Wordnet. The synonyms are least likely used by humans.

\subsubsection{Specificity \& Interestingness Reduction} A "Dullness" strategy is applied to make a response less specific and interesting by (1) Substituting the original response with a generic response. (2) Using a generic answer instead of the original answer in a question-answer pair. (3) Replacing the original response with one of the speaker's prior statements.

\subsection{Dialogue-Level Adversarial Perturbations}
\label{subsec:dialog-leve-adv-strategies}

The dialogue-level adversarial perturbations are motivated by prior works on self-supervised learning of automatic dialogue evaluation metrics~\citep{zhang-etal-2021-dynaeval,ghazarian-etal-2022-deam,zhang-etal-2022-fined}. Except for self-contradiction, we select 40 high-quality human-chatbot conversations from FED-Dial for each perturbation, and the perturbation is repeated five times per dialogue. Finally, we arrive at 200 unique data instances per perturbation. For self-contradiction, we rely on the existing human-written contradiction dataset, DECODE~\citep{nie-etal-2021-like}. 

\subsubsection{Coherence Reduction} Two adversarial strategies are applied to reduce the coherence level of a dialogue:

\smallskip
\noindent\textbf{(a). Order Shuffle} - The order of utterances in a coherent dialogue is randomized. 

\smallskip
\noindent\textbf{(b). Random Utterance Replacement} - An utterance in a coherent dialogue is replaced with random utterances drawn from other conversations. 

\subsubsection{Engagingness Reduction} Four strategies are implemented to reduce the engagingness level of a dialogue: 

\smallskip
\noindent\textbf{(a). Self-Contradiction} - Two human-written utterances are appended to a consistent dialogue such that one of the speakers makes a statement that contradicts one of their previous statements in the dialogue. 200 such consistent-contradiction pairs are randomly drawn from DECODE for metric robustness probing.

\smallskip
\noindent\textbf{(b). Utterance Duplication} - Given a high-quality human-chatbot dialogue, we prompt ChatGPT to rewrite the chatbot's utterances such that a portion of each chatbot utterance is appended to that utterance. In this way, the chatbot's utterances become unnatural.  

\smallskip
\noindent\textbf{(c). Repeating Others} - Given a high-quality human-chatbot conversation, we instruct ChatGPT to rephrase the chatbot's replies to consistently include portions of the human speaker's previous statements in its responses.

\smallskip
\noindent\textbf{(d). Generic Utterance} - We prompt ChatGPT to rephrase chatbot responses in a high-quality human-chatbot dialogue so that it always responds with generic statements.

\subsubsection{Informativeness Reduction} We adopt two strategies to reduce the informativeness and diversity of a dialogue. The first one is the ``\textbf{Generic Utterance}" strategy introduced previously and the second one is referred to as ``\textbf{Content Reduction}". Specifically, given a high-quality human-chatbot multi-turn dialogue, we prompt ChatGPT to simplify each chatbot response such that the information density in the response is low.

\begin{table*}[!ht]	
\centering
\resizebox{0.8\linewidth}{!}{
\begin{tabular}{@{}l|l@{}}
\toprule
\textbf{Perturbations} & \textbf{Descriptions}\\ \midrule
Random Response (RR) & \makecell[l]{The response is replaced with a response \\ from other dialogues or the dialogue history.} \\ \midrule
Replace Pronoun (RP) &  \makecell[l]{Pronoun usage in response is disturbed, such as  \\ swapping gender-specific pronouns with neutral ones.} \\ \midrule
Replace Named Entity (RNE) & \makecell[l]{Named entities in response are replaced \\ by other random entities.}\\ \midrule
Contradiction (Con) & \makecell[l]{The response's meaning is altered to contradict \\ the dialogue context.} \\ \midrule
Repetition (Rep) & \makecell[l]{A random portion of the response is repeated \\ multiple times to create make it unnatural.}\\ \midrule
Unnatural Paraphrase (UP) & \makecell[l]{Non-stop English words are randomly sampled and \\ replaced with unnatural synonyms from Wordnet.}\\ \midrule
Dullness (Dul) & \makecell[l]{The original response is replaced with a generic response \\ or one of the speaker's prior statements.}\\ \midrule
Order Shuffle (OS) & \makecell[l]{The order of utterances in a coherent dialogue is randomized.} \\  \midrule
Utterance Replacement (UR) & \makecell[l]{An utterance in a coherent dialogue is replaced with \\ a random utterance drawn from other conversations.} \\  \midrule
Self-Contradiction (SC) & \makecell[l]{Human-crafted utterances are added to a dialogue such \\ that one speaker contradicts their prior statements.}\\\midrule
Utterance Duplication (UD) & \makecell[l]{The chatbot's responses are modified by appending a segment of \\ the response back to itself, making the responses unnatural.}\\ \midrule
Repeating Others (RO) & \makecell[l]{The chatbot's responses are rephrased such that they consistently \\ include portions of the human speaker's previous statements.}\\ \midrule
Generic Utterance (GU) &\makecell[l]{ChatGPT is prompted to paraphrase chatbot replies in \\ dialogue, such that the chatbot always gives generic responses.}\\ \midrule
Content Reduction (CR) & \makecell[l]{We prompt ChatGPT to simplify the chatbot responses in a dialogue such \\ that the information density of the responses is low.}\\
\bottomrule
\end{tabular}}
\caption{The descriptions of different perturbation strategies.}
\label{tab:perturbation-descriptions}
\end{table*}
\end{document}